\documentclass{egpubl}
\usepackage{VMV2023}

\WsPaper           % uncomment for final version of EG Workshop contribution
\usepackage[T1]{fontenc}
\usepackage{dfadobe}  

\BibtexOrBiblatex
\electronicVersion
\PrintedOrElectronic

\ifpdf \usepackage[pdftex]{graphicx} \pdfcompresslevel=9
\else \usepackage[dvips]{graphicx} \fi

\usepackage{egweblnk} 
\usepackage[T1]{fontenc}
\usepackage{dfadobe}  
\usepackage{amsmath}

\usepackage{cite}  % comment out for biblatex with backend=biber
\usepackage{egweblnk} 
\usepackage{bbm}
\usepackage[table]{xcolor}% http://ctan.org/pkg/xcolor
\usepackage{hyperref}
\usepackage{algorithm}
\usepackage{algpseudocode}
\usepackage[normalem]{ulem}
\usepackage{dsfont}

\newcommand{\ADDITION}[1]{#1}
\newcommand{\REMOVAL}[1]{}
\newcommand{\CORRECTION}[2]{#2}

\DeclareMathOperator*{\argmax}{argmax}

\title{Improving NeRF Quality by Progressive Camera Placement for Free-Viewpoint Navigation}

\author[Georgios Kopanas \& George Drettakis]
{\parbox{\textwidth}{\centering Georgios Kopanas$^{1}$\orcid{0009-0002-5829-2192}
        and George Drettakis$^{1}$\orcid{0000-0002-9254-4819} }
 \\
{\parbox{\textwidth}{\centering $^1$ Inria \& Université Côte d’Azur, France}
}
}
\begin{document}

\teaser{
\includegraphics[width=\linewidth]{figures/teaser/teaser.ai}
\caption{
\label{fig:teaser}
We present a new method that proposes the next best camera placement for NeRF capture (left). We introduce two metrics that can be easily computed, observation frequency and angular uniformity (middle). On the right, we show that our approach outperforms two baseline camera placement strategies, \protect{\textsc{Hemisphere}} which is the typical approach used in most NeRF methods and \protect{\textsc{Random}}, as well as recent related work \protect{\cite{pan2022activenerf}}.
}
}

\maketitle
%-------------------------------------------------------------------------
\begin{abstract}
Neural Radiance Fields, or NeRFs, have drastically improved novel view synthesis and 3D reconstruction for rendering. NeRFs achieve impressive results on object-centric reconstructions, but the quality of novel view synthesis with free-viewpoint navigation in complex environments (rooms, houses, etc) is often problematic. While algorithmic improvements play an important role in the resulting quality of novel view synthesis, in this work, we show that because optimizing a NeRF is inherently a data-driven process, good quality data play a fundamental role in the final quality of the reconstruction. As a consequence, it is critical to choose the data samples -- in this case the cameras -- in a way that will eventually allow the optimization to converge to a solution that allows free-viewpoint navigation with good quality. Our main contribution is an algorithm that efficiently proposes new camera placements that improve visual quality with minimal assumptions. Our solution can be used with any NeRF model and outperforms baselines and similar work. 
%\GK{It can also be tuned by the user to focus on regions of interest to increase quality.}
   
%The tool at \url{http://dl.acm.org/ccs.cfm} can be used to generate
% CCS codes.
%Example:
\begin{CCSXML}
	<ccs2012>
	<concept>
	<concept_id>10010147.10010371</concept_id>
	<concept_desc>Computing methodologies~Computer graphics</concept_desc>
	<concept_significance>500</concept_significance>
	</concept>
	<concept>
	<concept_id>10010147.10010371.10010372</concept_id>
	<concept_desc>Computing methodologies~Rendering</concept_desc>
	<concept_significance>500</concept_significance>
	</concept>
	<concept>
	<concept_id>10010147.10010257.10010282.10011304</concept_id>
	<concept_desc>Computing methodologies~Active learning settings</concept_desc>
	<concept_significance>100</concept_significance>
	</concept>
	</ccs2012>
\end{CCSXML}

\ccsdesc[500]{Computing methodologies~Computer graphics}
\ccsdesc[500]{Computing methodologies~Rendering}
\ccsdesc[100]{Computing methodologies~Active learning settings}

\printccsdesc   
\end{abstract}  
%-------------------------------------------------------------------------
\section{Introduction}
%%\TODO{softer introduction}NeRF is having a big impact on novel view synthesis and 3D reconstruction for rendering.  Most notably, it is emerging as a better replacement to the complicated and involved optimization algorithms of Multi-View-Stereo\cite{mvs-schonberger2016} that are slow, and fail to produce faithful visual and geometric reconstruction. On the contrary, NeRFs\cite{mildenhall2021nerf} use Stochastic Gradient Descent (SGD) to fit a 3D differentiable volumetric representation on 2D captured images resulting in much better overall quality, with recent methods achieving high speed \cite{muller2022instant}\cite{fridovich2022plenoxels}. 
In recent years, Neural Radiance Fields (NeRFs)~\cite{tewari2022advances} have emerged as a powerful approach allowing high-quality novel view synthesis, for scenes captured with photos taken from many different viewpoints. 
These methods also provide an alternative to Multi-View-Stereo~\cite{mvs-schonberger2016} solutions that are slow and fail to produce faithful visual and geometric reconstruction. 
%%By using Stochastic Gradient Descent (SGD) to fit a 3D differentiable volumetric representation on 2D captured images, NeRFs provide much better overall quality, with recent methods achieving high speed \cite{muller2022instant}\cite{fridovich2022plenoxels}. 
%
For both MVS and NeRF, capturing a scene typically starts with users taking many photos or video of the scene. Usually, users follow instructions to loop around an object a few times at different heights and to make sure to also capture top views~\cite{luma-ai} which we call \emph{hemispherical} capture. This works well for ``object-centric scenes'', i.e., scenes that have a main object that the users want to be able to view freely, while the rest is considered background.
There has been little previous work on how to capture more general scenes such as rooms, buildings etc. that have no central point of interest, especially when the goal is to allow \emph{free-viewpoint} navigation in the environment. Users typically place the cameras based on their intuition and empirical knowledge about which camera placements usually work, often leading to the failure of the reconstruction and consequently, forcing users to recapture the scenes in a costly and time consuming trial-an-error process.

%\begin{itemize}
%\item Intuitively, hemispherical capture works for object-centric scenes because it samples the space containing the object uniformly in positions and the angular domain
%\item important to allow multi-view information to disambiguate depth
%\item Achieving such uniform coverage both in positions and angles is much more challenging in the context of general scenes
%\item With an infinite number of cameras, this could theoretically be achieved, but in practice the number of cameras is limited
%\item in addition, limitations in space
%\item The problem we try to solve is given a camera budget, how can we choose the next camera that will allow the distribution to be as close as uniform in space and angle
%\end{itemize}

Intuitively, hemispherical capture works for object-centric scenes because it samples the space containing the object \emph{uniformly} both for camera positions and in the angular domain. This uniform coverage is a dense sampling of a complete radiance field, since rays from the camera centers through the pixels in each view frustum cover space uniformly. Such coverage provides multi-view information that is used in the optimization to disambiguate depth, allowing accurate reconstruction.

Achieving such uniform ray coverage both in positions and angles is much more challenging in the context of general scenes, where there is no single central object. 
With an infinite number of cameras, it could theoretically be possible to densely sample the light field, but in practice the number of cameras is limited and in addition given the geometry of the scene cameras cannot be placed everywhere (i.e., inside objects).
The problem we try to solve is: given a camera budget and physical limitations of space, how can we \emph{efficiently} choose the next camera that will allow the resulting ray sampling to be as close as possible to uniform in space and angle.

%\begin{itemize}
%\item In this paper we first develop an metric to evaluate uniformity in space and angle that is fast to evaluate.
%\item We then propose an algorithm that uses this metric to select the next best camera that will make the overall distribution closer to uniform
%\item We evaluate the metric and the algorithm on synthetic data and compare to baselines and previous work
%\item We also run our algorithm on a real dataset as a proof of concept
%\end{itemize}

There has been little previous work on this problem; most methods that have been proposed require either modification on the training of the NeRF model making them unsuitable for generalization to other NeRF variants other than the ones that it was specifically designed for, or very expensive calculations based on the current state of the NeRF model. These properties make the process slow and cumbersome.

In our solution, we first develop a metric to evaluate uniformity in space and angle that is fast to evaluate.
We then propose an algorithm that uses this metric to select the next best camera such that the overall distribution will be closer to uniform in positions and angle.
We evaluate the metric and the algorithm on synthetic data and compare to baselines and previous work, demonstrating that our solution works well, and we also run our algorithm on a real dataset as a proof of concept.
%\begin{itemize}
%\item From a practical perspective, our algorithm could be used in the future for automated capture using robotic or drone capture; we leave this as future work, but discuss a practical future use case in Sec.~\ref{sec:use-case}.
%\end{itemize}
%
From a practical perspective, our algorithm can be used for automated capture using robotic or drone capture; we leave this as future work, but we discuss a practical future use case in Sec.~\ref{sec:use-case}.

In summary, our contributions are:
\begin{itemize}
\item The definition of an efficient metric for \emph{observation frequency} and \emph{angular uniformity} that can be computed on the fly during NeRF capture, without requiring additional images.
\item An algorithm to quickly estimate reconstruction quality of a scene and that proposes the next camera placement that maximizes the improvement in quality of capture based on our metrics.
\end{itemize}
	
Our solution can work with any NeRF model without changes to the optimization loop, and only introduces a small performance overhead to the training. We performed extensive testing on synthetic scenes, and our method achieves the best quality against multiple baselines and other algorithms that we tried given a limited budget of cameras. We also present a first preliminary evaluation with real data, in which our method also performs well. %\TODO{talk about perfomance in results?} \TODO{missing references everywhere}

%-------------------------------------------------------------------------

\section{Related Work}
\label{sec:related}

In recent years a huge number of publications on Neural Radiance Fields (NeRFs) have been published; we will start by reviewing only a few papers that are closely related to our method and design choices.  Recent comprehensive surveys on NeRFs can be found in~\cite{tewari2022advances} and~\cite{xie2022neural}.
We then review camera selection for \CORRECTION{recontruction}{reconstruction}, both traditional and for NeRFs. % and briefly discuss some examples of active learning that are related to our approach.
%-------------------------------------------------------------------------
\subsection{Neural Radiance Field Basics}
%-------------------------------------------------------------------------
Neural Radiance Fields were introduced by \cite{mildenhall2021nerf}. They fundamentally changed how we can reconstruct 3D scenes from 2D images by introducing a continuous volumetric representation of the scene, encoded in a Multi-Layer Perceptron (MLP) which we can optimize using Stochastic Gradient Descent (SGD) to fit the input images, solving the reconstruction with a data-driven optimization.

Follow-up work improved the reconstruction quality by allowing for better extrapolation when dealing with cameras observing objects from different distances \cite{barron2021mip} but also generalized the algorithms for unbounded and realistic scenes \cite{barron2022mip}. 
Most of the results demonstrated in these papers focus on object-centric camera placements which has become the ``typical NeRF-style capture'', i.e., a hemisphere of cameras around the object and looking directly at the center of the object.

Another line of work focuses on performance, both for fast training and fast rendering. Most solutions achieve good results by encoding the radiance field in voxel grids with limited spatial extent\cite{muller2022instant,fridovich2022plenoxels,sun2022direct,chen2022tensorf} or point clouds\cite{xu2022point}. For our experiments we will built on top of Instant-NGP~\cite{muller2022instant}, since it is currently the NeRF method with the best quality/performance trade-off. Instant-NGP uses a hash function to map a 3D hierarchical voxel-grid of high dimensional features to a compact 1D representation. This grid is later queried in an optimized way along the ray to produce the final color for each pixel. The optimization includes a very efficient \emph{occupancy grid} that marks the voxels as occupied or free and results in an efficient way to skip empty space during ray marching.

\ADDITION{Also, there are numerous papers that try to reconstruct NeRFs from a very limited amount of input views\cite{yu2021pixelnerf,jain2021putting, kulhanek2022viewformer}. These models could potentially be benefited greatly by an optimal selection of cameras.}

While all these algorithms significantly improve the state-of-the-art,
in the vast majority of cases they use datasets in which the cameras are placed on a hemisphere over a region of interest. This allows for good quality reconstruction only in that specific region (typically an object of interest). It is not clear how one would place cameras for more complicated environments, when allowing the user to navigate freely. Camera placement is an important factor that controls the final quality of the reconstruction and the ability to \emph{navigate freely} in the scene without artifacts.
In this context we propose a solution that will automate and standardize the way of capturing NeRFs, removing the burden of trial and error from the user.

\subsection{Camera placement for reconstruction}

We discuss representative previous work in camera placement for reconstruction. In the vast majority of cases we assume that the scene is captured with photographs, and that the cameras of these photographs have been calibrated. Camera calibration is typically performed using Structure-from-Motion (SfM), using systems such as COLMAP~\cite{colmap}.

\subsubsection{Traditional Reconstruction}
Multi-View Stereo (MVS) is an offline process that recovers the 3D geometry from a set of images and is very computationally expensive. The user first captures the images, performs camera calibration and then runs MVS. After a few hours of computation, the user may come to realize that the images are not good enough for a good 3D reconstruction, requiring the scene to be recaptured from scratch. This is a tedious process, especially if accessing the capture site is difficult.

To improve this cumbersome process the field of Next-Best-View (NBV) estimation~\cite{isler2016information,dunn2009next,bircher2016receding}
%\TODO{CHECK ADD OTHERS} 
predicts the next view that will provide more information to the reconstruction process given a set of already captured views. In the field of volumetric reconstruction \cite{isler2016information} focuses on sensors with depth and creates a set of heuristics to estimate the next best view that will maximize the information gain of each newly acquired sample. While this work is inspiring it lies outside the scope of optimizing a model from a set of data-samples with SGD because they use a depth sensor that directly observes the geometry information of the scene, while in neural radiance fields the geometry representation is being optimized to fit the scene.

Other works in camera selection that focus on MVS can be separated in heurstic-based methods~\cite{mostegel2016uav,smith2018aerial} or data-based~\cite{liu2022learning}. They mostly focus on  estimating or predicting the uncertainty of the MVS reconstruction process without actually running it. 
%Please note that 
In contrast to MVS, fast NeRF models~\cite{muller2022instant,fridovich2022plenoxels} open the door for new approach in the field of next best view estimation that allows online reconstruction and camera placement prediction, especially if camera calibration can be provided online by the capture device (e.g., augmented reality helmet).

%-------------------------------------------------------------------------
\subsubsection{Neural Radiance Fields}
%\TODO{same title as above, kinda confusing}
Automatic camera placement for Radiance Fields is an emerging topic of research. A popular approach is to modify the NeRF model to be able to predict it's own uncertainty \cite{ran2023neurar,zeng2022efficient,pan2022activenerf} which later is used in various ways to choose the views which maximize it. The uncertainty is modeled in two ways, either by converting the MLP that encodes the scene to a Bayesian MLP \cite{ran2023neurar,zeng2022efficient,pan2022activenerf} that also predicts it's own uncertainty or by using the physical properties of the volumetric representation along a ray based on the entropy of the density function\cite{zhan2022activermap}. All methods that use the NeRF model to predict uncertainty are computationally intensive since they need many MLP evaluations for each candidate camera. In addition, it is hard to train an MLP that predicts it's own uncertainty. That is why \cite{ran2023neurar} uses a depth sensor to stabilize the training. Some methods \cite{pan2022activenerf} focus on selecting views when there is a very limited budget of cameras allowed and  \cite{lee2022uncertainty} presents a solution that evaluates the uncertainty based on the spread of density along a ray. This needs a full rendering step per candidate camera which means when the space of candidates grows in unconstrained environments \ADDITION{it comes} with an increased cost. All the above methods are not demonstrated on non object-centric scenes, making them unsuitable for our context which focuses on free-viewpoint navigation in complex scenes.

\section{Method}

The goal of our method is to dynamically suggest new camera positions 
such that we create a dataset that will achieve a good quality reconstruction. This can be used to guide a robotic agent or a human to acquire new images when capturing a NeRF.
NeRFs trained on object-centric datasets achieve excellent quality when observing the object from a camera that matches the distribution of the training cameras, but easily break when moving away from them, see Fig.~\ref{fig:comparisons_ours}. We are interested in constructing a carefully designed placement of cameras that will allow the final user of the NeRF to navigate freely in the scene, while avoiding 
%breaking the virtual immersion by 
strong visual artifacts.

We want to generalize the simple assumptions of the object-centric capture style to more complex scenes and viewing scenarios, in particular when we allow the viewer to navigate freely.

\subsection{Observation Frequency and Angular Uniformity}
The object-centric capture style of NeRF\cite{mildenhall2021nerf} and MipNeRF\cite{barron2021mip} has two main properties. First, all cameras observe the object and second, the cameras are distributed along different directions to cover the angular domain uniformly. 
If we constrain the user to view the scene on the hemisphere, this capture style naturally provides a good reconstruction since it 
%fits the assumptions of a good dataset as it 
covers the space all the possible cameras uniformly. 

We next provide a formal definition of this observation, and in particular a measure of \emph{observation frequency} and a measure of \emph{angular uniformity} of these observations.

Given a set of cameras $\mathcal{C}$, a point $p$ in space will be well reconstructed if it is observed often from the input cameras and if these cameras are distributed uniformly in the angular domain of directions. We next formalize this mathematically and generalize it for multiple points $p$.

We define a function that describes how frequently each \REMOVAL{is} point is observed. For a point $p$ we define the frequency $O_f(p)$ of observation as follows:

\begin{equation}\label{eq:freq_cam}
	O_f(p) = \frac{\sum_{i=0}^{N}{\mathds{1}_{obs}(\mathcal{C}_i, p)}}{N}
\end{equation}

Where $\mathds{1}_{obs}$ is an indicator function that is $1$ if point $p$ lies inside the frustum  of camera $C_i$ and $N$ is the total number of cameras. This equation describes a simple relationship between cameras and points in space: If all the available cameras observe a point, then $O_f(p)=1$, while if no cameras observe it $O_f(p)=0$.

\begin{figure}[!t]
	\includegraphics[width=\linewidth]{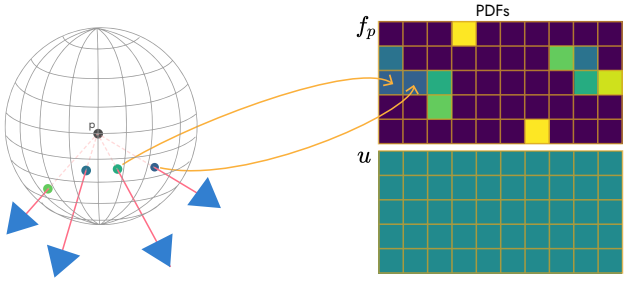}
	\caption{
		To compute the distance of the empirical distribution in the angular domain of cameras that observe a point $p$, we assign the cameras to bins based on their direction from point $p$ into a histogram in polar coordinates. We then convert this to a PDF, for which we account for the non-uniform surface area of the spherical coordinate system. Then we use the total variation distance in Eq.(\ref{eq:wasser}) to get the value for node point $p$.
	}
	\label{fig:pdf_wasser}
\end{figure}

For the directions of the observations, we next define a metric to measure \emph{angular uniformity}, since more uniform angular distribution of observed directions results in better resulting visual quality.

We define $f_p$ the distribution of camera directions in the angular domain that observe a point $p$ and the uniform distribution $u$ in the same angular domain. To determine the quality of the angular distribution of cameras,
we will compute the total variation distance between the two distributions.

\begin{equation} \label{eq:wasser}
	TV(f_p, u) = \frac{1}{2}\sum_{\theta,\phi \in \Omega}{| f_p(\theta,\phi) - u(\theta, \phi) |}
\end{equation}

where \CORRECTION{$F_p$ and $U$ are the corresponding CDFs}{$u$ is a uniform PDF}. We construct the piece-wise constant PDF $f^C_p$ in the angular domain by computing the histogram of the directions of the cameras that observe point $p$. Every bin in the histogram contains the number of cameras that observe this point from the solid angle that corresponds to the bin. 
%The Wasserstein-1 metric, can be seen as the \TODO{VERIF difference of the} areas \TODO{of} \sout{between} the empirical CDFs $F_p$ and the uniform CDF $U$ \TODO{area isnt really defined; what do you mean ? is there a formal term ? }. 
Similarly to Eq.~(\ref{eq:freq_cam}), the angular metric is 0 if our point p is observed from a uniform distribution, while it approaches 1 as it moves further from the uniform distribution. 

We provide a visual illustration of this process in Fig.\ref{fig:pdf_wasser}, showing the histogram of $F_p$ and $U$ in bins with polar coordinates. The direction are represented in polar coordinates, that not area preserving, so we weight the bins of the histogram based on the surface area of each bin.

\subsection{Estimating Reconstruction Quality}

We define the area where we want to estimate the reconstruction quality, and for simplicity we use an axis-aligned bounding box to define it. This is the area which the user wishes to observe; We refer to this area as $\mathcal{B}$. 

Ideally, we would like to evaluate the quality of reconstruction of every point in $\mathcal{B}$. In practice however, we discretize the problem by constructing a regular grid in $\mathcal{B}$ with resolution of $32^3$ and refer to it as $\mathcal{B}_{32}$; we will evaluate reconstructability on the \emph{nodes} of the grid. In discrete space it is easier to measure the total reconstructability $E$ of $\mathcal{B}$ given by the set of cameras $\mathcal{C}$ by summing over all the nodes $p$ of the grid:

\begin{equation}\label{eq:energy}
	E(\mathcal{C}, \mathcal{B}_{32}) = \sum_{p\in \mathcal{B}_{32}}{(1 - TV(f_p^\mathcal{C}, u)) + O_f(p)^{\gamma}},
\end{equation}

\noindent
where $f_p^\mathcal{C}$ is the empirical PDF in the angular domain for point $p$ and set of cameras $\mathcal{C}$ and $u$ is the uniform PDF in the same domain,
%\TODO{re-defined them above, remove?}, 
while $\gamma$ is a non linear scaling factor that modulates how much more important it is to observe points that are less frequently observed than points that have already been observed frequently.

Our formulation has several advantages. First, it is easily interpretable by a user, making it easy to modify and specify regions that have more importance than others. In the limit case of a single point and if $C$ is constrained on a hemisphere our method reduces to the typical ``object-centric'' capture setup of NeRF and MipNeRF/360. Second, a key advantage is that our camera proposal does not require additional image acquisition to estimate reconstruction quality, as for other methods, e.g., based on uncertainty estimation~\cite{pan2022activenerf}. Our combined observation frequency and angular uniformity metrics can be seen as a proxy for uncertainty, while being relatively cheap to compute.

\subsection{Optimization}
\label{sec:optim}

Our goal is to find the set of cameras $\mathcal{C}$ that maximize the quantity in Eq.~(\ref{eq:energy}):
\begin{equation}
	\argmax_{\mathcal{C}} E(\mathcal{C}, \mathcal{B}_{32})
\end{equation}

To do this we optimize the NeRF model while we choose the new cameras. Cameras can only be placed in empty space and some NeRF models, including Instant-NGP \cite{muller2022instant} that we use, provide an occupancy grid that is used to skip empty space while rendering. If the implementation does not provide one it is trivial to compute it on the fly by sampling 3-D space and storing the density \ADDITION{and using a threshold to binarize the value}. We use the occupancy grid to place candidate cameras in free space. % and also to evaluate Eq.~(\ref{eq:energy}) in the subset of the nodes in $\mathcal{B}_{32}$ that lie in occupied space. \TODO{the subset thing is out of the blue, maybe skip it discuss it elswhere or omit it if we dont have it at results.} 
For simplicity, we also constrain the cameras to lie inside $\mathcal{B}$.

\begin{algorithm}[!h]
	\caption{Summary of the greedy optimization algorithm to minimize Eq.~(\ref{eq:energy}})\label{alg:optimization}
	\begin{algorithmic}
		\State $T \gets 100$ \Comment{Total number of cameras we want to sample.}
		\State $N \gets 1000$ \Comment{Number of cameras we sample in each iteration.}
		\State $nodes \gets \mathcal{B}_{32}$
		\State $\mathcal{C} \gets [\hspace{0.1cm}]$
		\While{$ |\mathcal{C}| <= T$}
			\State $aabb \gets  \mathcal{B} \hspace{0.2cm} \textbf{if} \hspace{0.2cm} |\mathcal{C}| > 20 \hspace{0.2cm} \textbf{else} \hspace{0.2cm} \mathcal{B}^f$
			\State $C_p \gets sampleRandomCameras(N, aabb) $
			\ForAll {$c \in \mathcal C_p $}
				\State $ E_c \gets E(\mathcal{C}\cup c, nodes)$ \hspace{0.4cm} Eq.~(\ref{eq:energy})
			\EndFor
			\State $c_{max} \gets \argmax_{c} E_c$
			\State $C\text{.insert}(AcquireImage(c_{max}))$
			\If{$|\mathcal{C}| > 20$}
			\State $train\_nerf(iterations=250)$
			\EndIf
		\EndWhile
	\end{algorithmic}
\end{algorithm}
In the beginning of our run we have no images nor a trained occupancy grid, and one needs the other to initialize the process. To overcome this, we ask the user to create a bounding box in the scene that is empty and call it $\mathcal{B}^f$. This allows us to sample cameras safely so we can start the process. This is somewhat of a chicken-and-egg problem, since we need enough of an initial reconstruction to have a coordinate system to define this initial box $\mathcal{B}^f$. In a real-world scenario, the user will simply take 20 photos of the scene -- evidently in free space -- that will initialize the reconstruction and the coordinate system, and imply the definition of $\mathcal{B}^f$. 
For the synthetic examples we used for evaluation, we have the coordinate system beforehand, and we defined $\mathcal{B}^f$ manually.

Now that we know how to sample random cameras safely, we want to maximize the quantity in Eq.~(\ref{eq:energy}). We use a greedy maximization technique: every 250 \ADDITION{training} iterations we acquire a new camera given a set of already chosen $\mathcal{C}_k$. We do this by sampling a set of N=1000 candidate cameras that lie inside $\mathcal{B}$, or inside $\mathcal{B}^f$ for the first 20 cameras. Each camera gets a random direction, and we filter out all cameras whose center lies in occupied space or observes occupied space from too close. Then we compute $E(\mathcal{C}_k \cup c_n, \mathcal{B}_{32})$ for each of the N cameras, choose the camera with the highest score and add it in $\mathcal{C}_{k+1}=\mathcal{C}_k\cup c_n$. We repeat until we acquire as many cameras as our budget allows. 
The algorithm is summarized in Alg.~\ref{alg:optimization}.
Our current unoptimized implementation requires a few seconds to propose a new camera; further optimization would allow truly interactive use.
%Note that this process is fast, around 2-3 sec. per iteration for our unoptimized implementation and is thus useable for interactive capture. \TODO{VALIDATE THE NUMBERS}

\subsection{Future Practical Usage Scenario}
\label{sec:use-case}
The method above provides all the elements for online camera selection that can be used in future work by a robotic or drone-based system. In this paragraph we describe how such a system could function to motivate the utility of our results.

As discussed above, a user would first take 10-20 photos of the environment for an initial camera calibration, providing a reference frame, and allowing the definition of an initial box $\mathcal{B}^f$. In a fully operational system, we would then run a fast NeRF such as InstantNGP~\cite{muller2022instant} that reconstructs a first approximation of the scene volumetric representation in a few seconds. We then would run our algorithm to choose the next camera, and move the robot or drone to the next position; the new capture is then incrementally added to the NeRF optimization until the budget of cameras is reached. Given the latency of moving the capture agent and the fact that we need a few training iterations between the two captures, a slightly optimized implementation of our algorithm is perfectly suited to such a scenario, since it can provide the next best camera faster than the agent can actually move to the next position. As a consequence, a full system using our algorithm would allow automatic and high quality capture with a small number of photos, both reducing capture time and optimizing resulting image quality.

\begin{table*}[!ht]
	\caption{
		\label{tab:comparisons-per-scene} {Per Scene Quantitaive evaluation of our method. We provide the PSNR for each test set separately and the total average of each algorithm for each scene.}
	}
	\small
	\scalebox{0.67}{
		\begin{tabular}{l|cccc|cccc|cccc|cccc|cccc|}
			
			Scene & \multicolumn{4}{c|}{\textsc{LivingRoom1}}  & \multicolumn{4}{c|}{\textsc{LivingRoom2}} & \multicolumn{4}{c|}{\textsc{Office6}} & \multicolumn{4}{c|}{\textsc{Office9}} & \multicolumn{4}{c|}{\textsc{Kitchen5}} \\
			Train|Test
			&Random&Hsphere&Ours& Avg
			&Random   & Hsphere & Ours & Avg
			& Random   & Hsphere & Ours & Avg
			& Random   & Hsphere & Ours & Avg
			& Random   & Hsphere & Ours & Avg\\
			\hline 
			Hsphere    & 17.43 & \textbf{31.48} & 14.63 & 21.18 & 19.13 & \textbf{30.12} & 13.04 & 20.76 & 20.34 & \textbf{30.48} & 15.22 & 22.01 & 15.17 & \textbf{36.79} & 13.23 & 21.73 & 18.92 & \textbf{33.28} & 16.36 & 22.85 \\
			ActiveNerf & 20.57 & 17.91 & 18.59 & 19.02 & 25.51 & 22.76 & 21.28 & 23.18 & 24.71 & 20.78 & 19.65 & 21.71 & 25.75 & 27.52 & 24.40 & 25.89 & 25.00 & 22.83 & 21.42 & 23.08 \\
			Random     & 25.77 & 23.86 & 22.81 & 24.14 & 26.65 & 24.48 & 22.00 & 24.37 & 28.57 & 25.63 & 22.06 & 25.42 & 26.20 & 28.29 & 22.65 & 25.71 & 25.24 & 24.30 & 22.70 & 24.08 \\
			Ours       & \textbf{27.46} & 27.80 & \textbf{26.86} & \textbf{27.37} & \textbf{28.40} & 27.02 & \textbf{26.02} & \textbf{27.14} & \textbf{31.35} & 27.01 & \textbf{27.12} & \textbf{28.49} & \textbf{28.38} & 30.59 & \textbf{28.91} & \textbf{29.29} & \textbf{27.74} & 27.06 & \textbf{25.87} & \textbf{26.89} \\
			
		\end{tabular}
	}
\end{table*}

\begin{figure*}[!h]
	\setlength{\tabcolsep}{0.8pt}
	\centering
	\begin{tabular}{ccccc}
		GT&Ours&ActiveNerf\cite{pan2022activenerf}& Random&Hemisphere \\
		\includegraphics[width=.19\linewidth]{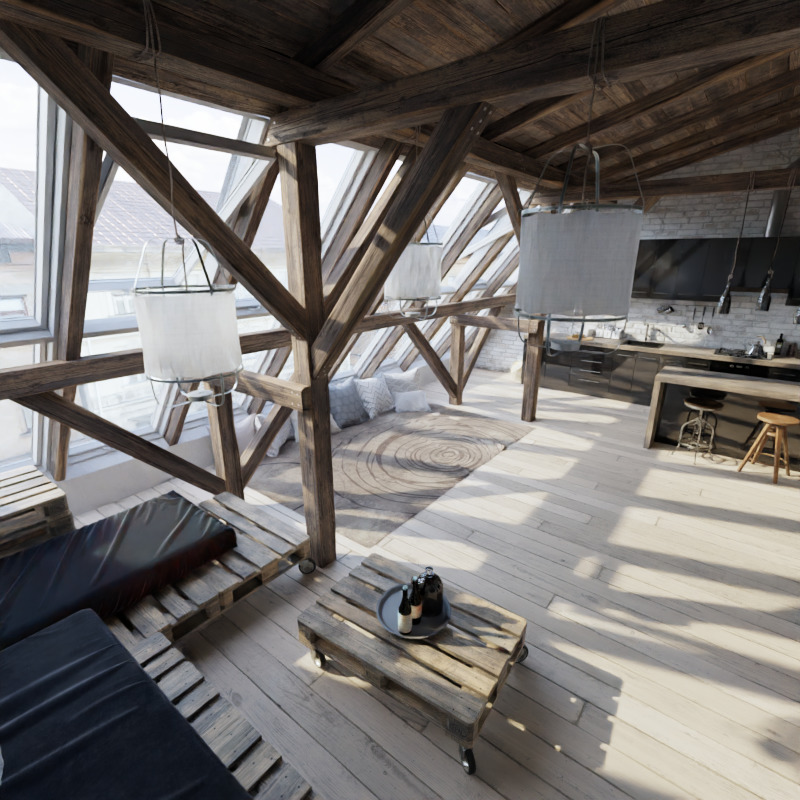} &
		\includegraphics[width=.19\linewidth]{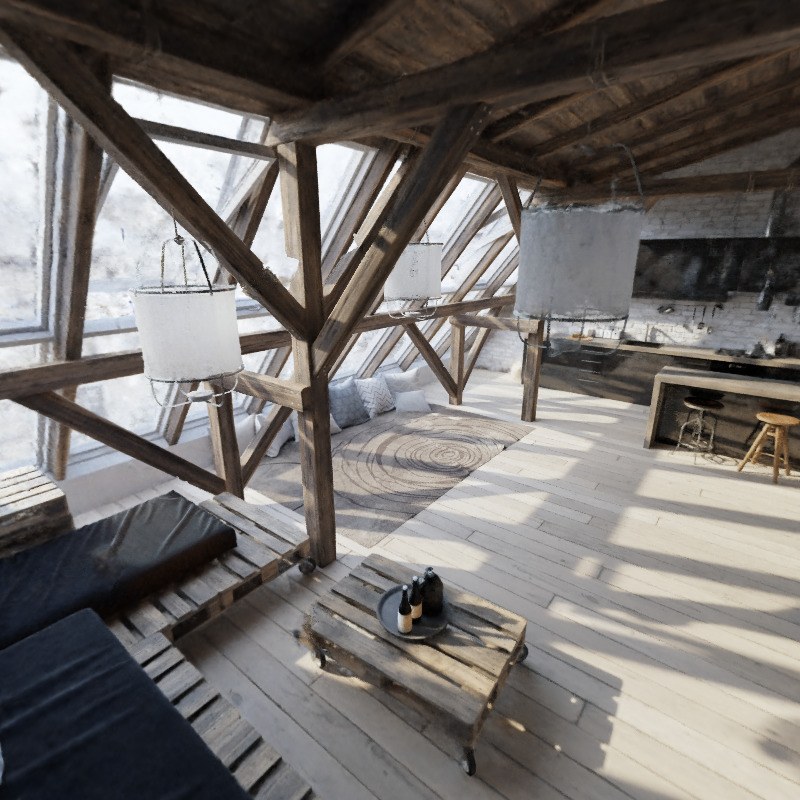} &
		\includegraphics[width=.19\linewidth]{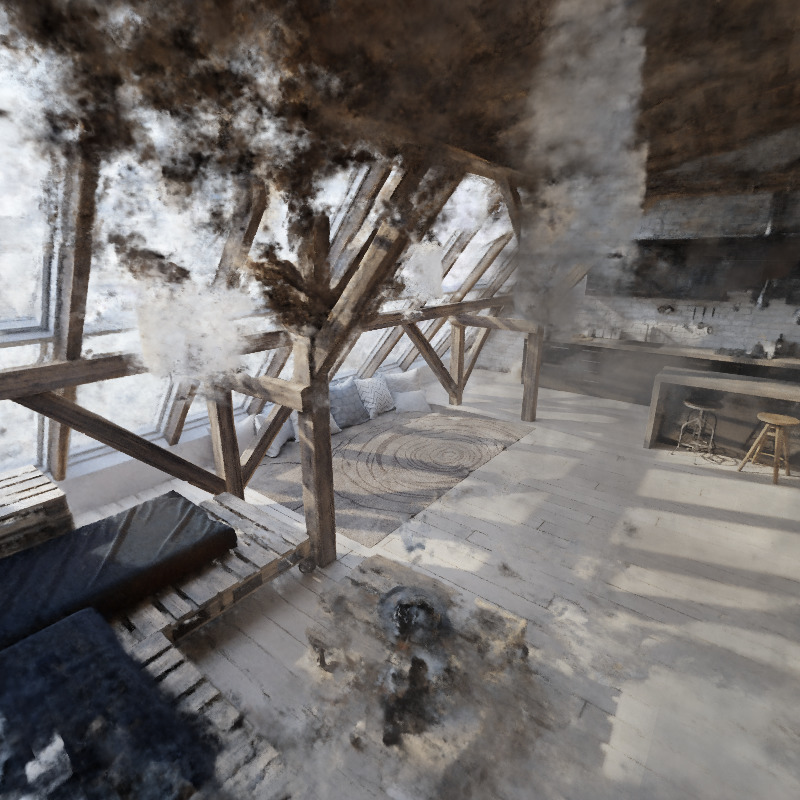} &
		\includegraphics[width=.19\linewidth]{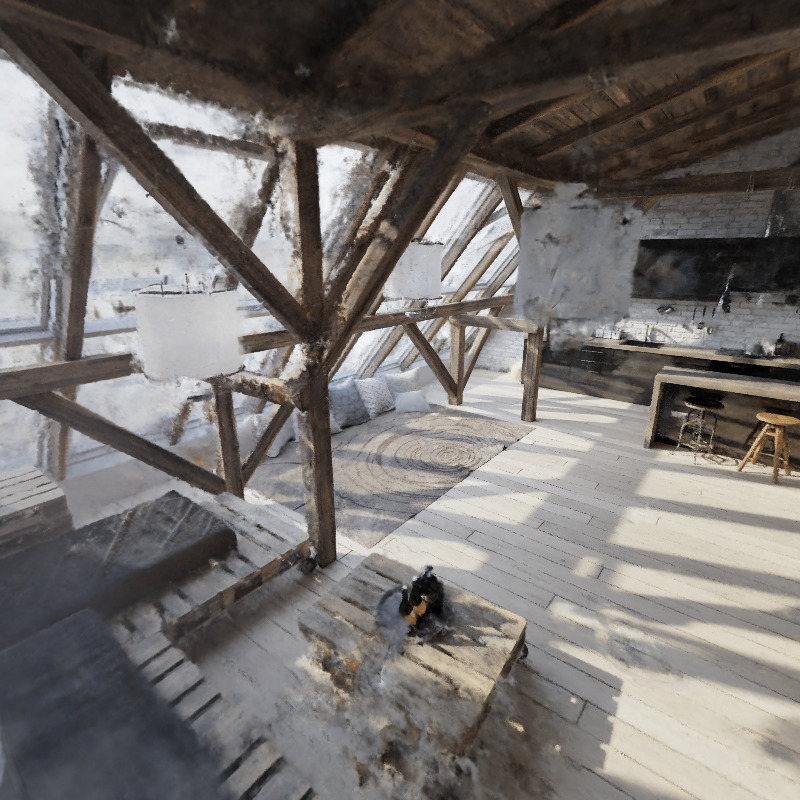} &
		\includegraphics[width=.19\linewidth]{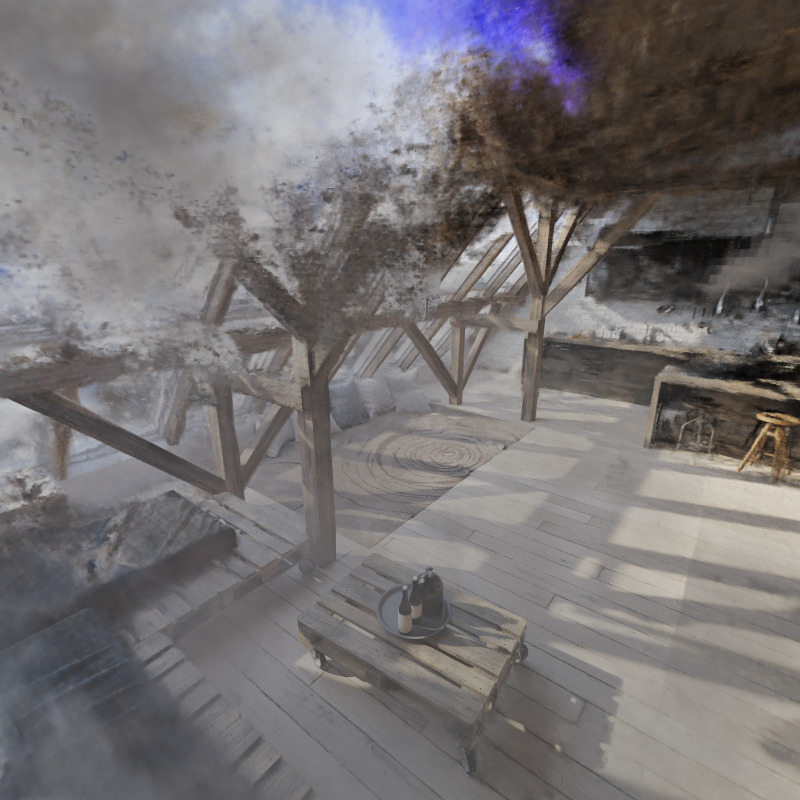} \\
		\includegraphics[width=.19\linewidth]{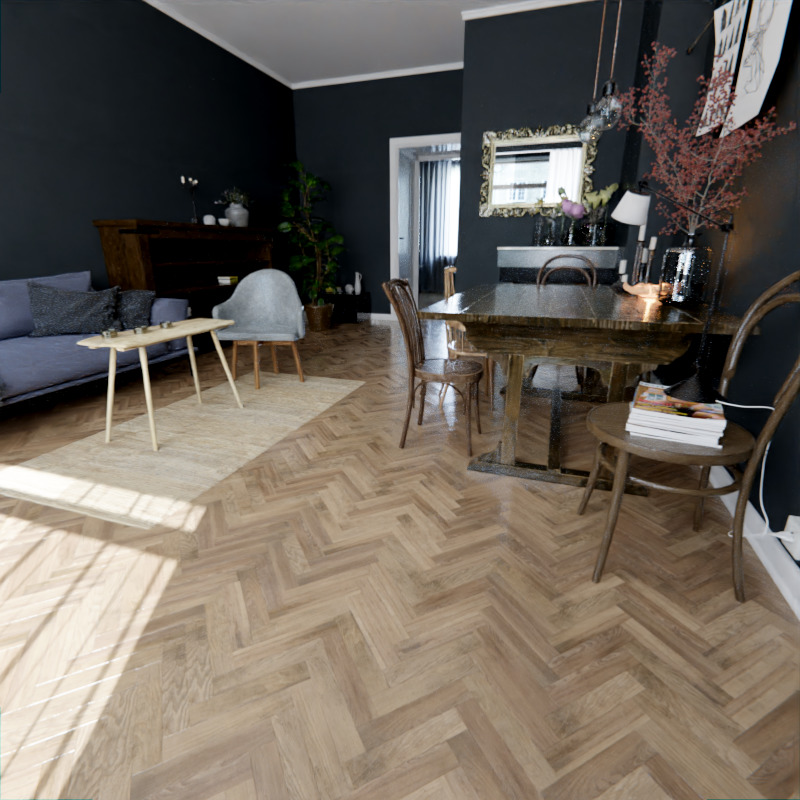} &
		\includegraphics[width=.19\linewidth]{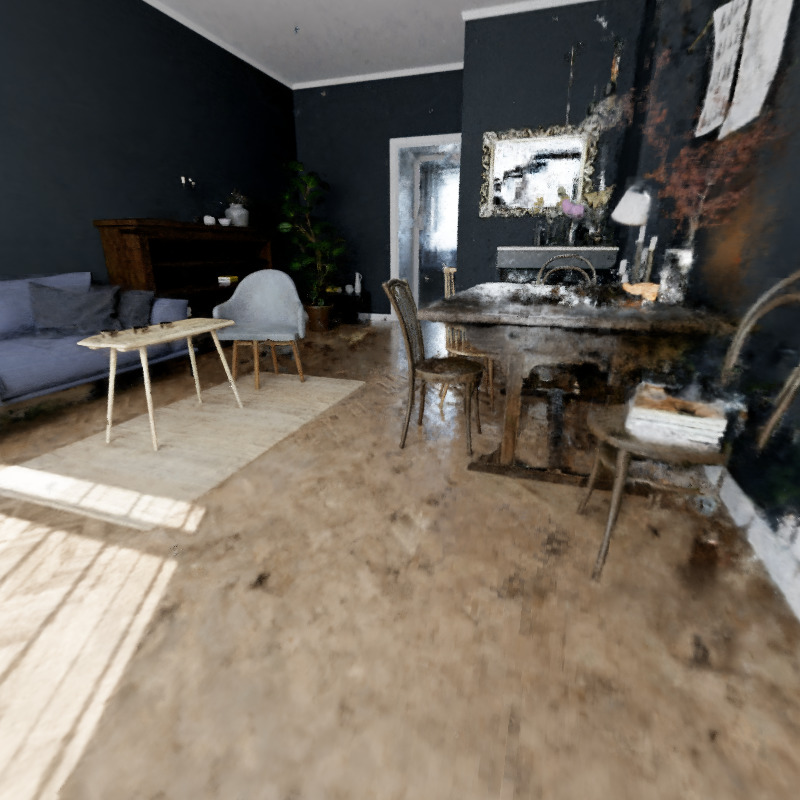} &
		\includegraphics[width=.19\linewidth]{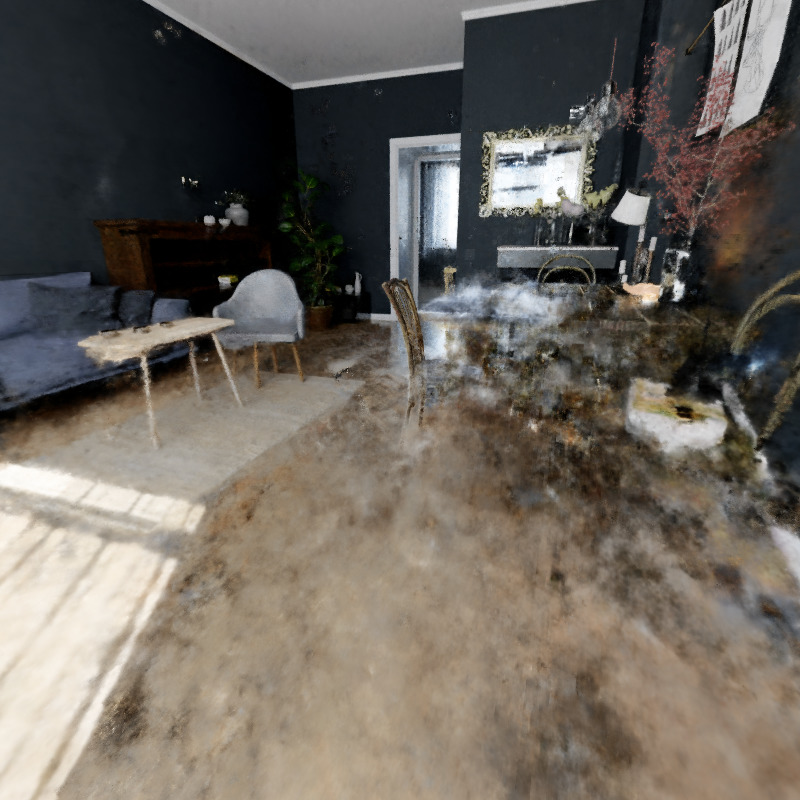} &
		\includegraphics[width=.19\linewidth]{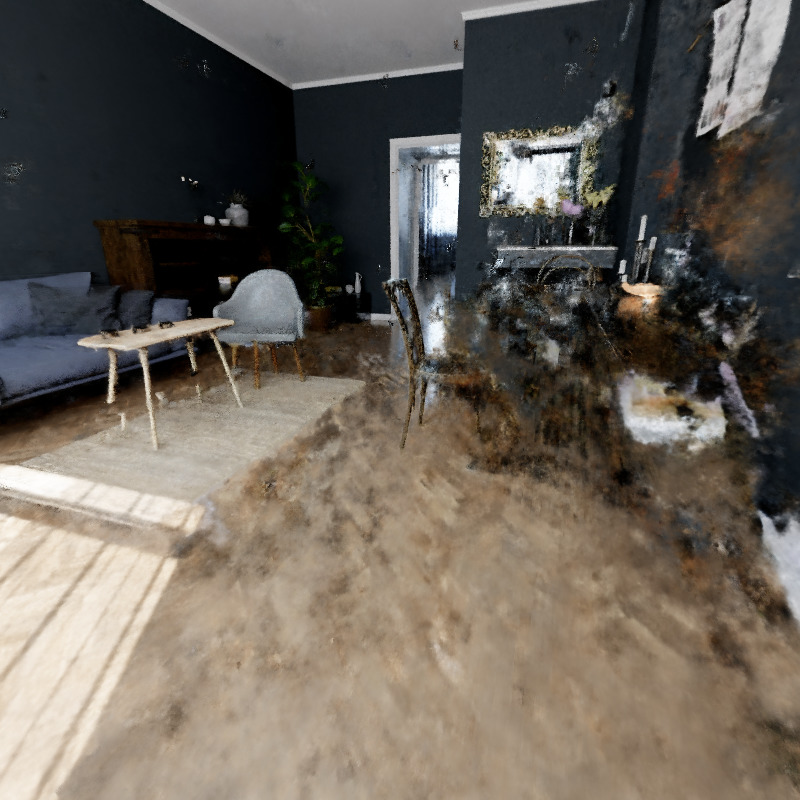} &
		\includegraphics[width=.19\linewidth]{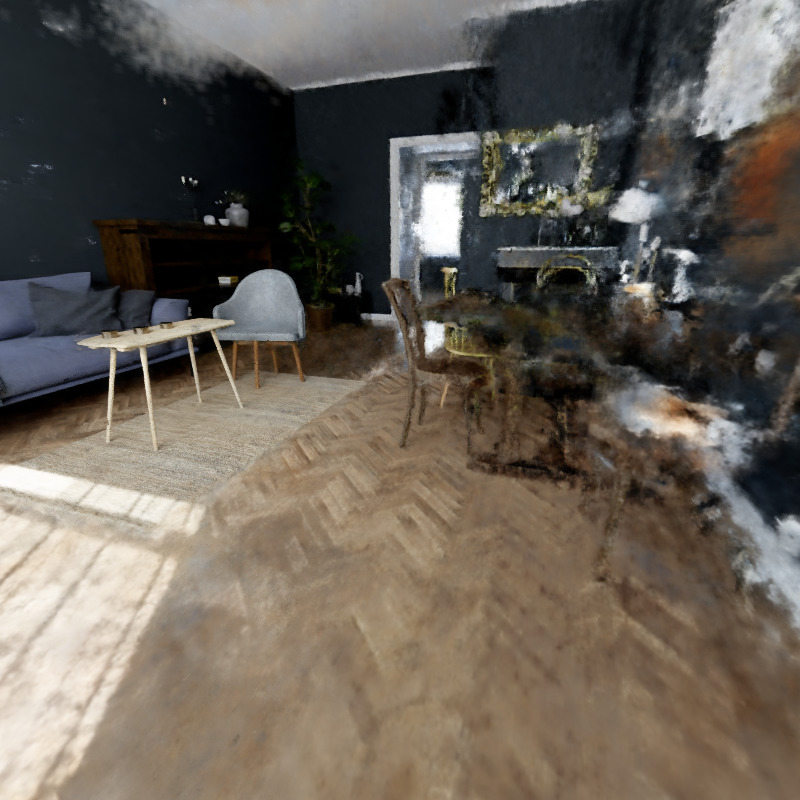} \\
		\includegraphics[width=.19\linewidth]{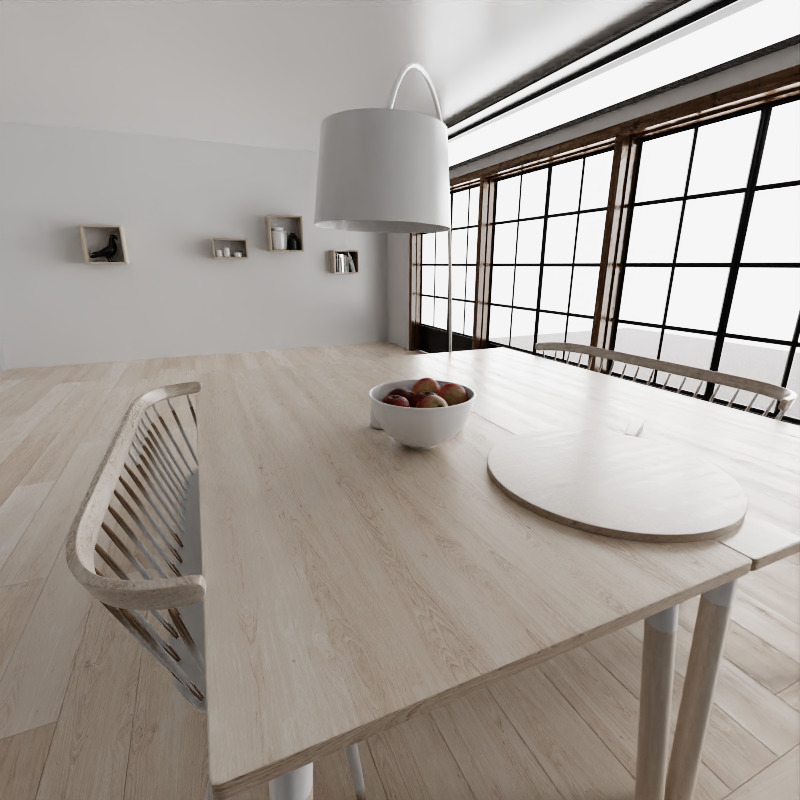} &
		\includegraphics[width=.19\linewidth]{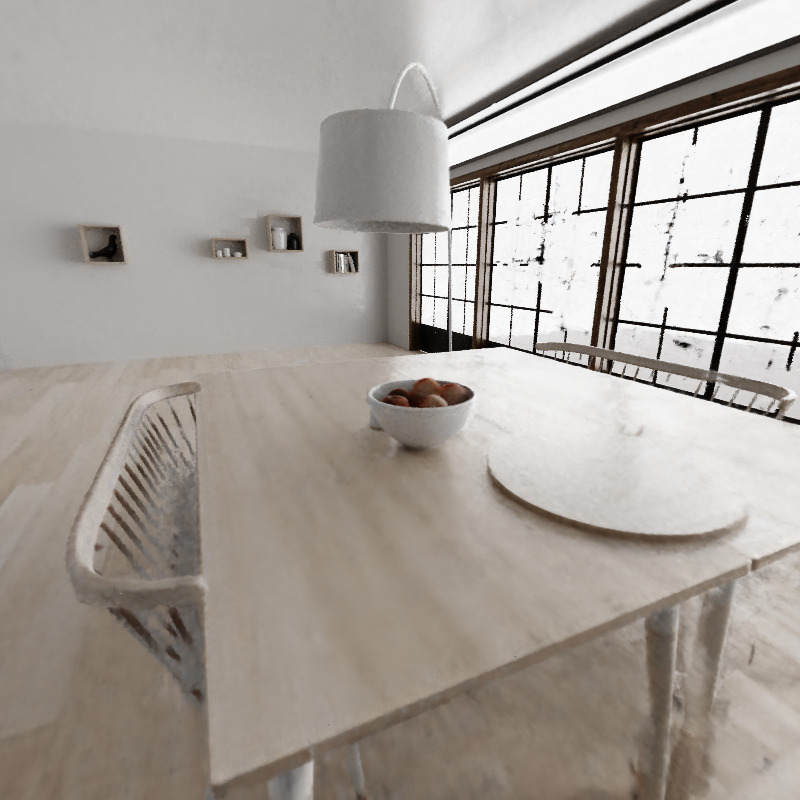} &
		\includegraphics[width=.19\linewidth]{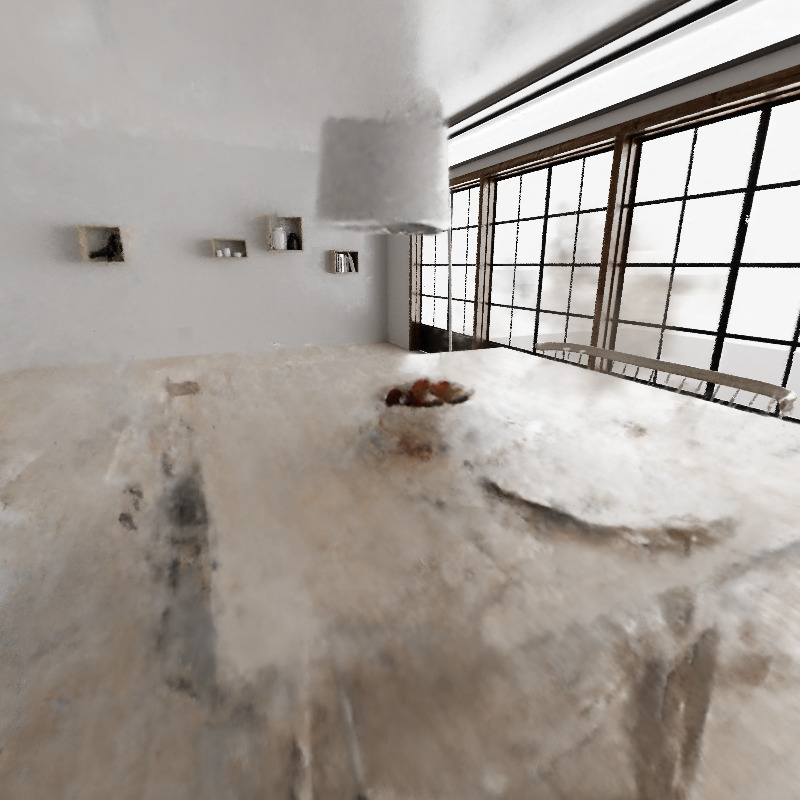} &
		\includegraphics[width=.19\linewidth]{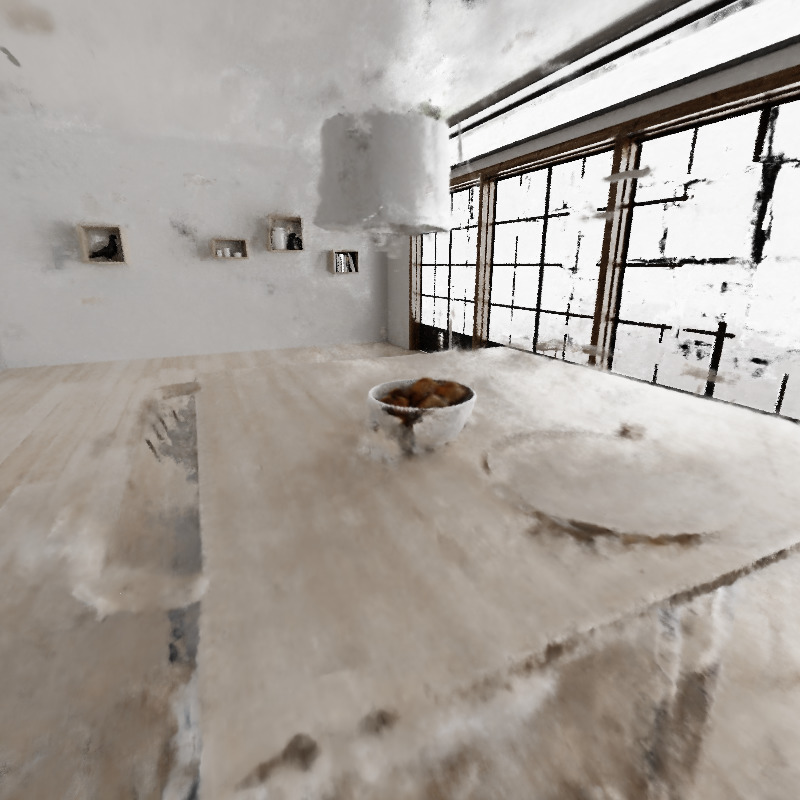} &
		\includegraphics[width=.19\linewidth]{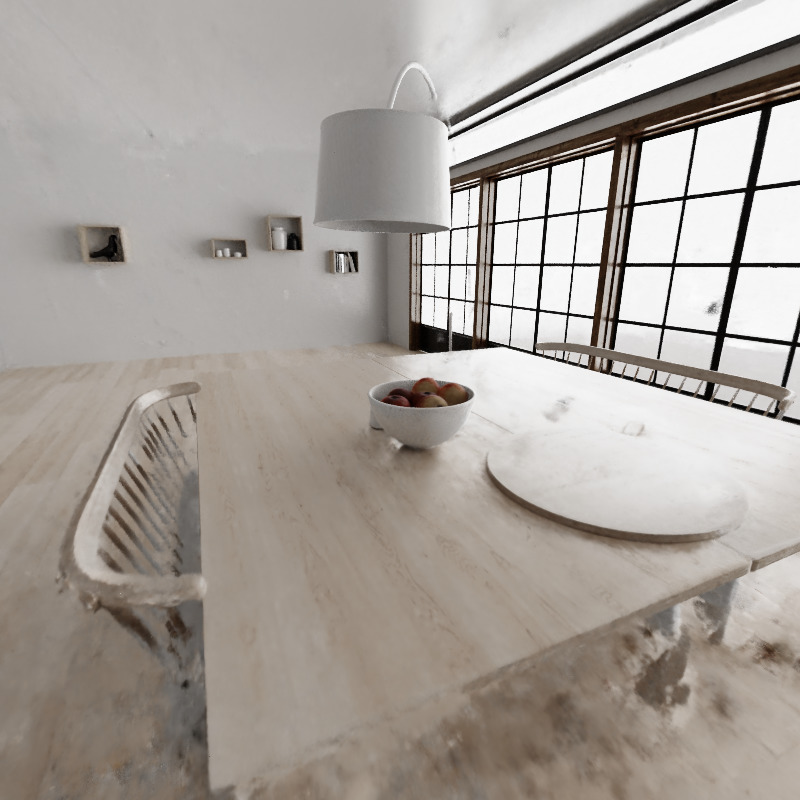} \\
		\includegraphics[width=.19\linewidth]{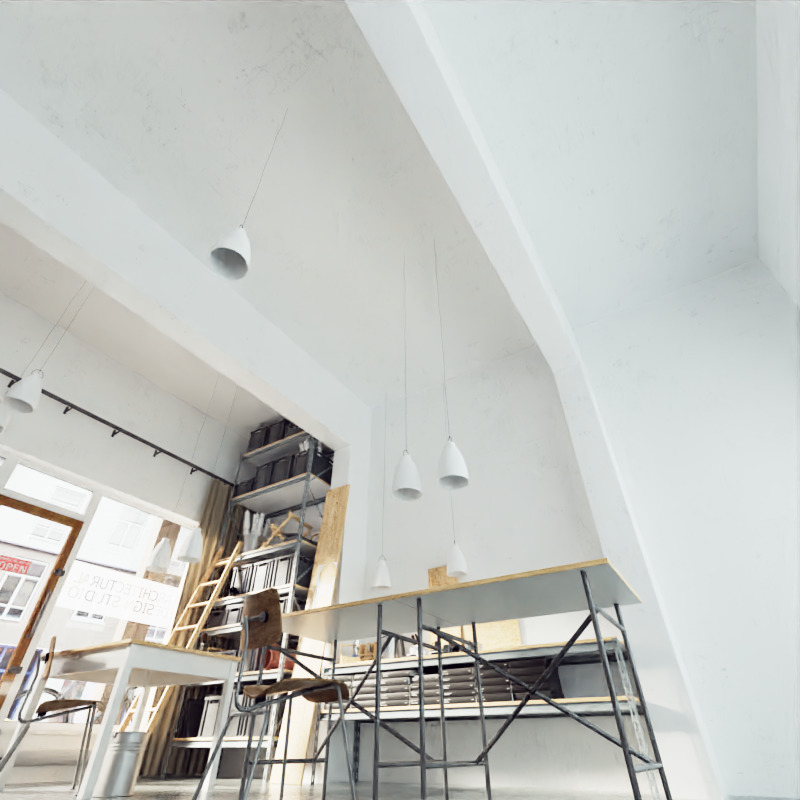} &
		\includegraphics[width=.19\linewidth]{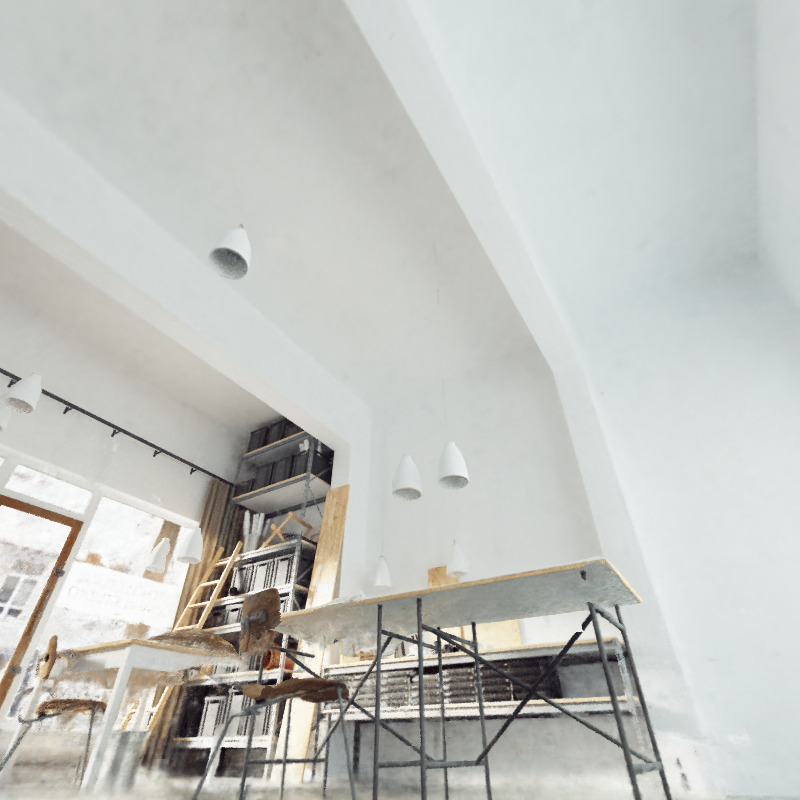} &
		\includegraphics[width=.19\linewidth]{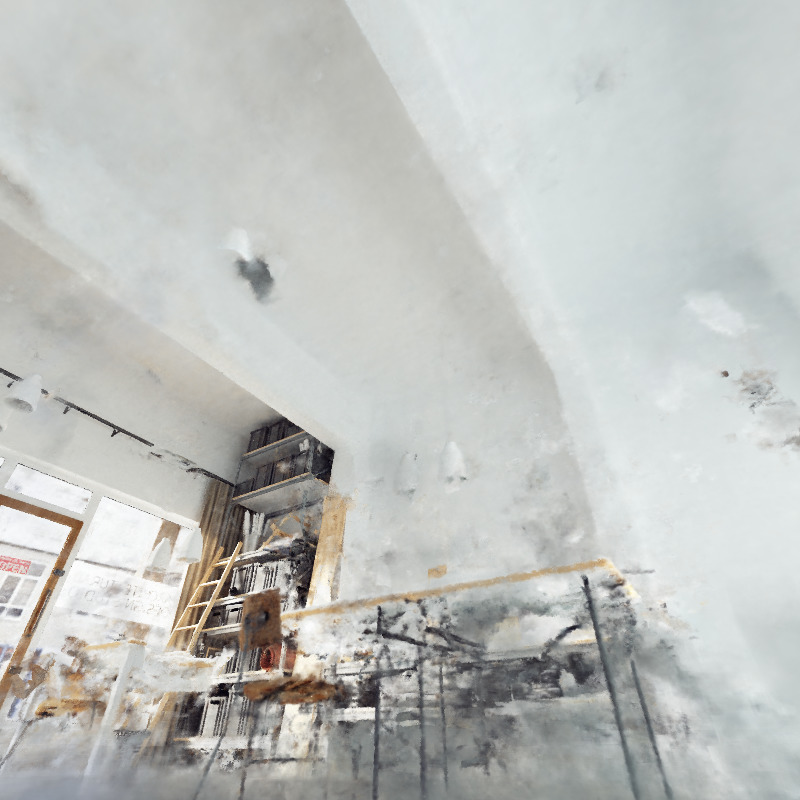} &
		\includegraphics[width=.19\linewidth]{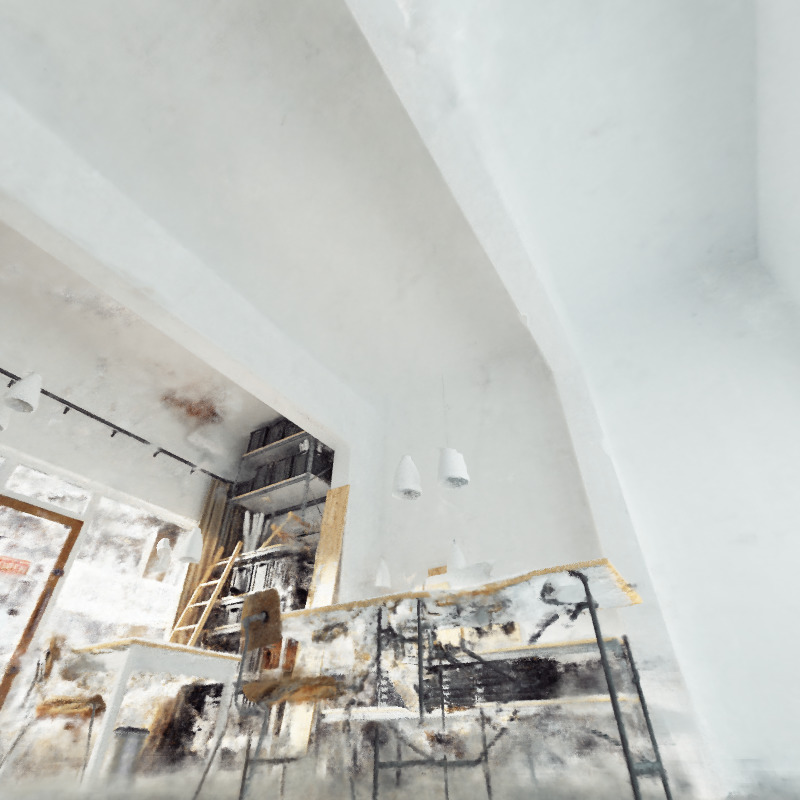} &
		\includegraphics[width=.19\linewidth]{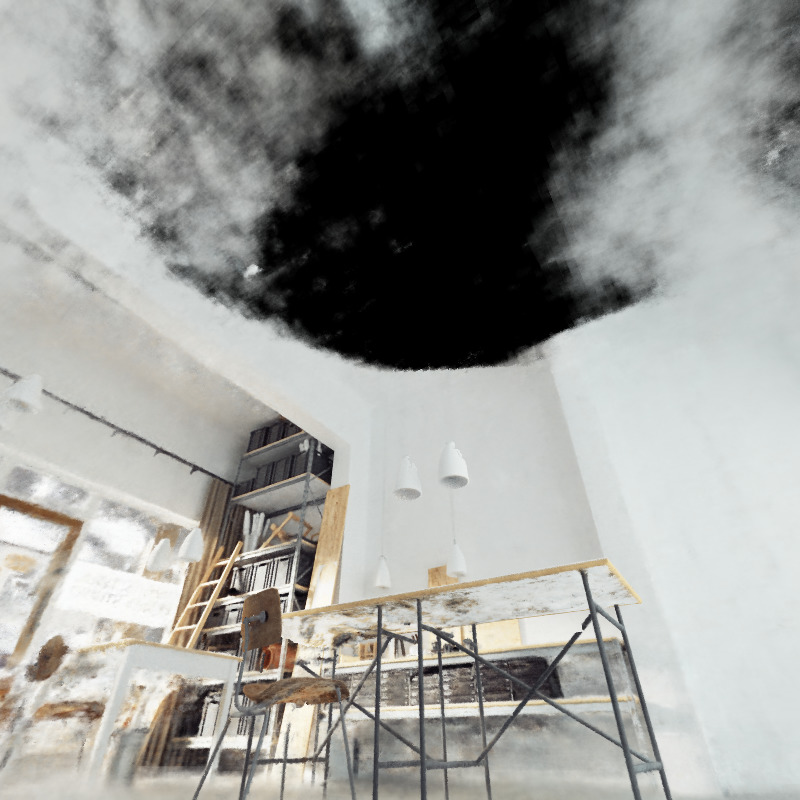} \\				\includegraphics[width=.19\linewidth]{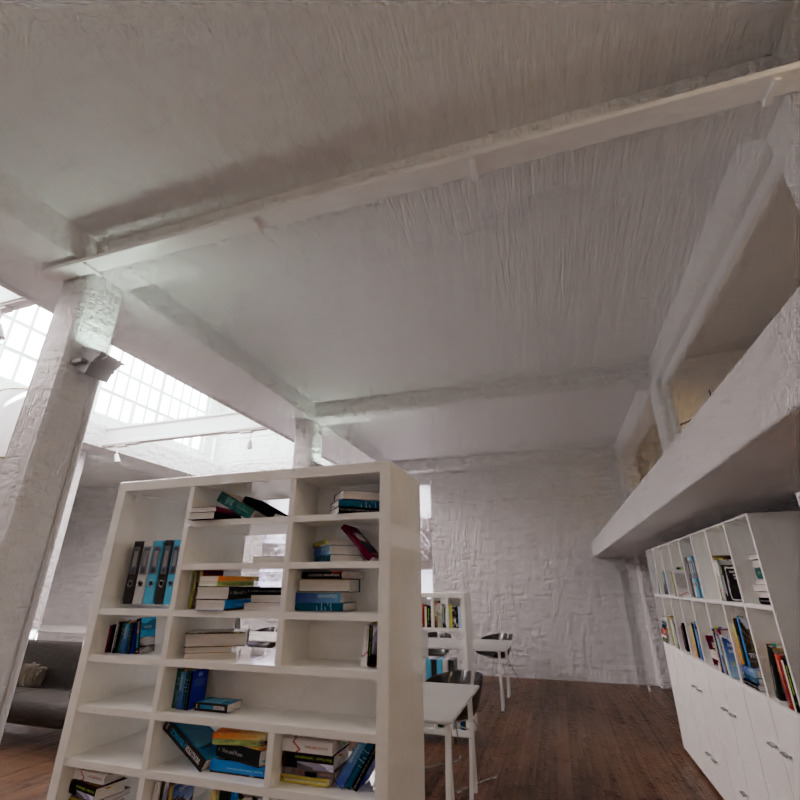} &
		\includegraphics[width=.19\linewidth]{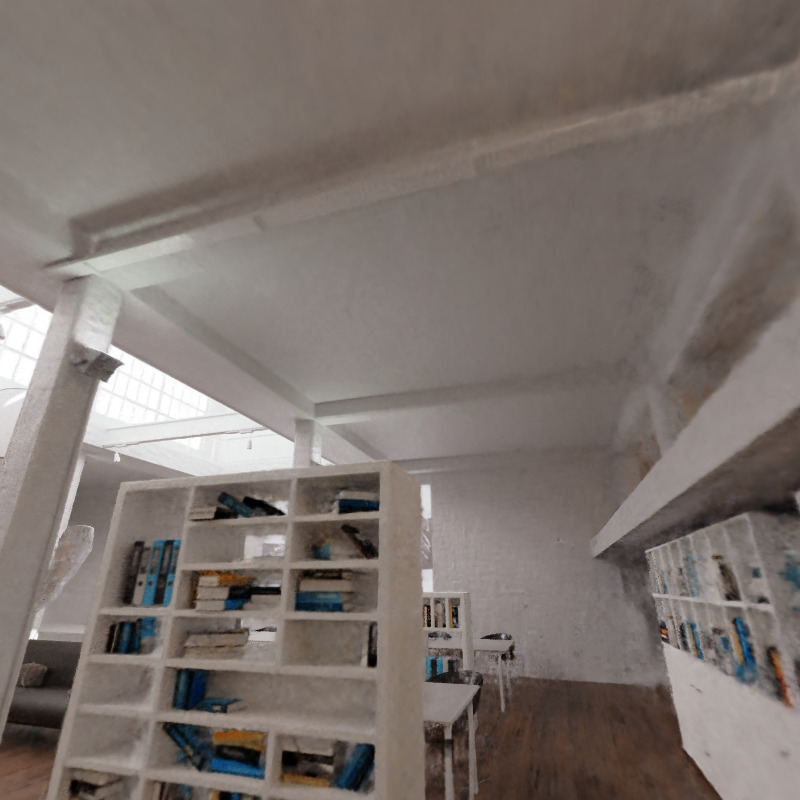} &
		\includegraphics[width=.19\linewidth]{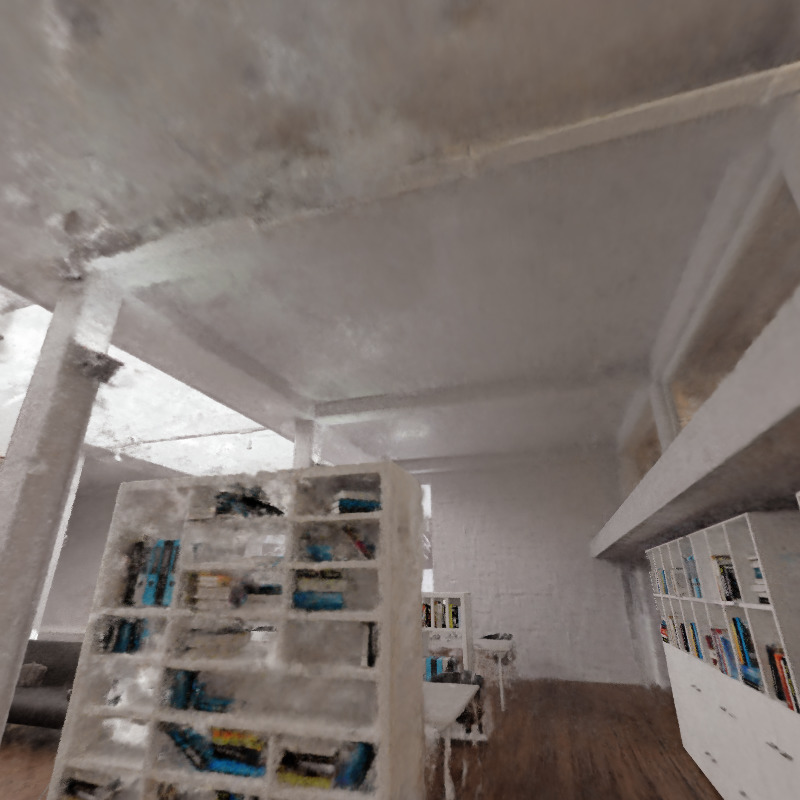} &
		\includegraphics[width=.19\linewidth]{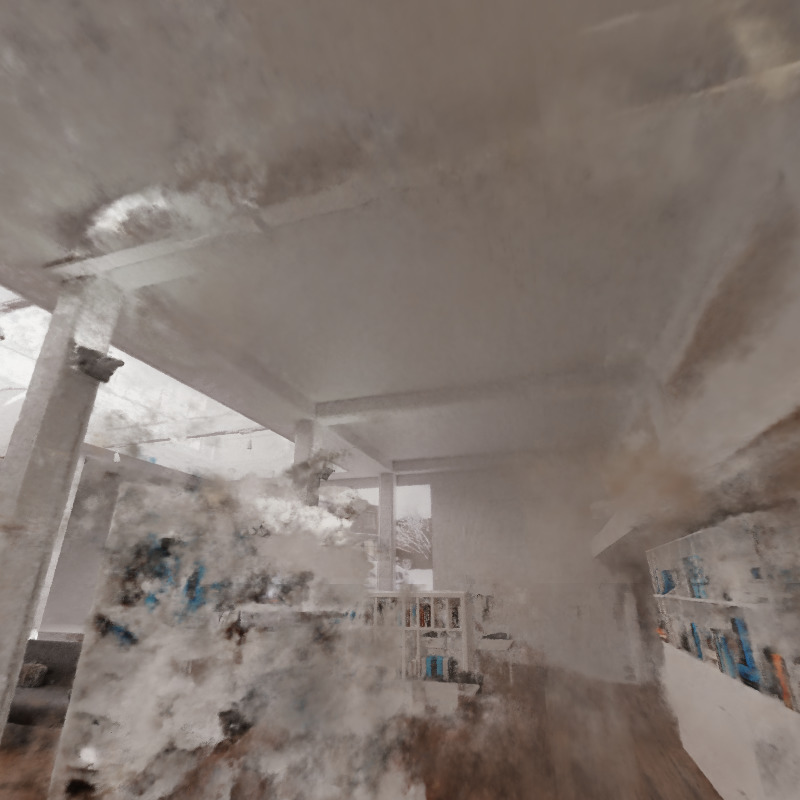} &
		\includegraphics[width=.19\linewidth]{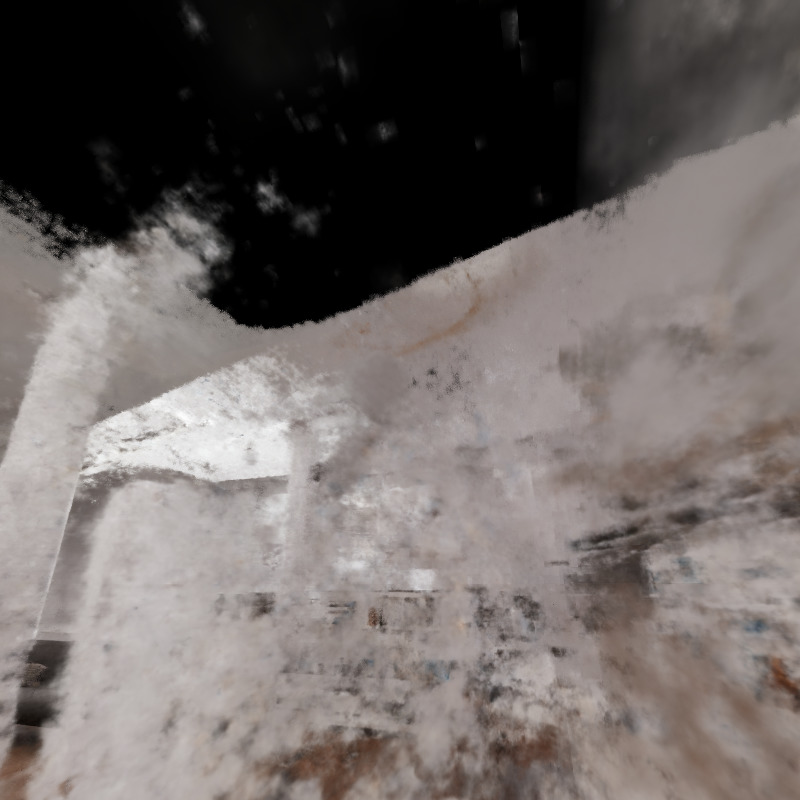} \\
	\end{tabular}
	\caption{
		\label{fig:comparisons_ours}
		Images from our sampling test set. We present a visual comparison to baselines(\textsc{Random},\textsc{Hemisphere}) and ActiveNerf. The first column shows the ground truth image. The scenes shown are \textsc{Livingroom1}, \textsc{Livingroom2}, \textsc{kitchen5}, \textsc{office6} and \textsc{office9}.
	}
\end{figure*}

%-------------------------------------------------------------------------
\section{Results \& Comparisons}

Evidently, the goal of our method is to provide guidance for a human or robotic acquisition system while capturing a NeRF. Designing the user interface for human guidance or interfacing with an automated acquisition system are complex tasks that we leave for future work. Instead, we provide a thorough evaluation of synthetic scenes, in which image acquisition is achieved simply by rendering a new image from the camera proposed by our system. We also provide a preliminary evaluation of our method on a real capture, by capturing a large number of views that we can then sample. %\TODO{VERIF THAT WE DO THIS}

This system was implemented by interfacing together Instant-NGP~\cite{muller2022instant} with Blender's python API \cite{blender} and Cycles renderer\footnote{https://www.cycles-renderer.org/}. We extended the python interface of Instant-NGP to allow us to query the occupancy grid efficiently and we also linked Instant-NGP with Blender's python environment. We will provide all source code and data, that will be available here \url{https://gitlab.inria.fr/fungraph/progressive-camera-placement}.
%TODO{next sentence has syntax problems}
The full pipeline that allows the treatment of complex synthetic scenes interfaced with NeRF systems such as InstantNGP is a powerful tool in itself, and was very useful for this project. We hope it will also be helpful to others experimenting with NeRFs in full, realistic scenes.

\begin{figure}[!h]
	\includegraphics[width=\linewidth]{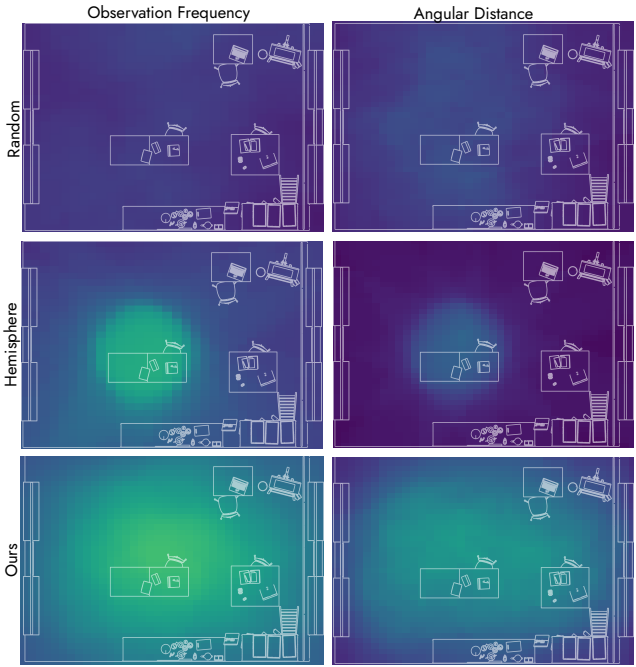}
	\caption{
		``Floorplan'' visualisation of the seperate elements that construct our energy term in Eq.~(\ref{eq:energy}). The values have been averaged along the Y axis.
	}
	\label{fig:floorplan_plot}
\end{figure}

\subsection{Evaluation on Synthetic Scenes}

For the first set of synthetic scene comparisons, we evaluate our method against two baseline camera placement strategies: \textsc{Hemisphere} where we place the cameras on a hemisphere around an object of interest\footnote{We manually select a point and a radius for the hemisphere such that it makes for that specific scene configuration, i.e a table with vases etc} in the scene (this is the standard NeRF\cite{mildenhall2021nerf} capture style) and \textsc{Random} where we place the cameras at a random position and a random orientation by also making sure that the camera placement is not in occupied space (i.e., not inside objects), see also Sec.~\ref{sec:optim}. 

\begin{figure}[!h]
	\includegraphics[width=\linewidth]{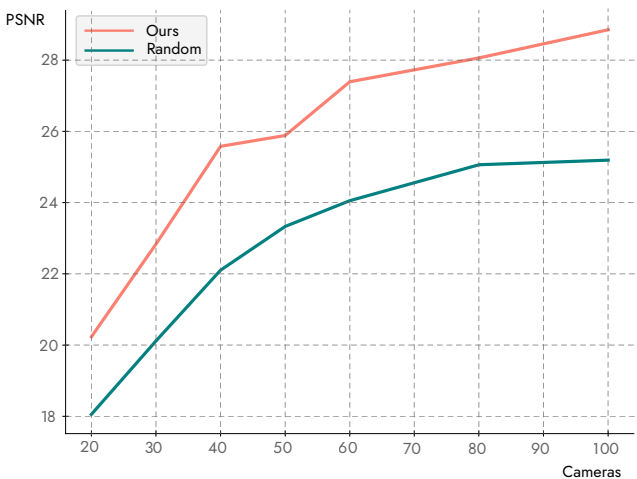}
	\vspace*{-5mm}\caption{ In this plot we show the test-set PSNR with relation to the number of cameras in the training set. We plot two sampling algorithms, random and ours. We evaluate the PSNR across all 150 images of the test set for the scene Office6 }
	\label{fig:train_line}
\end{figure}

As discussed in Sec.~\ref{sec:related}, there are few methods that treat our specific problem; of all NeRF camera selection methods, only ActiveNerf~\cite{pan2022activenerf} provides code, and we thus include a comparison to this approach.
We use the authors' implementation which is based on an implementation of the original NeRF~\cite{mildenhall2021nerf} method. To allow a best-effort fair comparison we used their code to extract the cameras, and then trained the same Instant-NGP\cite{muller2022instant} model as we did with all other baselines and our method. To extract the cameras we pre-render 1000 random cameras which act as the pool from which ActiveNeRF can choose cameras. The ActiveNeRF implementation is very computationally and memory intensive so we tuned ActiveNerf to start from 20 random cameras, and choose 8 cameras every 50k iterations. These 8 cameras are the best cameras chosen from their algorithm from a random subset of 100 out of the 1000 cameras. 

\begin{figure*}[!h]
	\setlength{\tabcolsep}{0.8pt}
	\centering
	\begin{tabular}{ccccc}
		GT&Ours&ActiveNerf\cite{pan2022activenerf}& \textsc{Random}&\textsc{Hemisphere} \\
		\includegraphics[width=.19\linewidth]{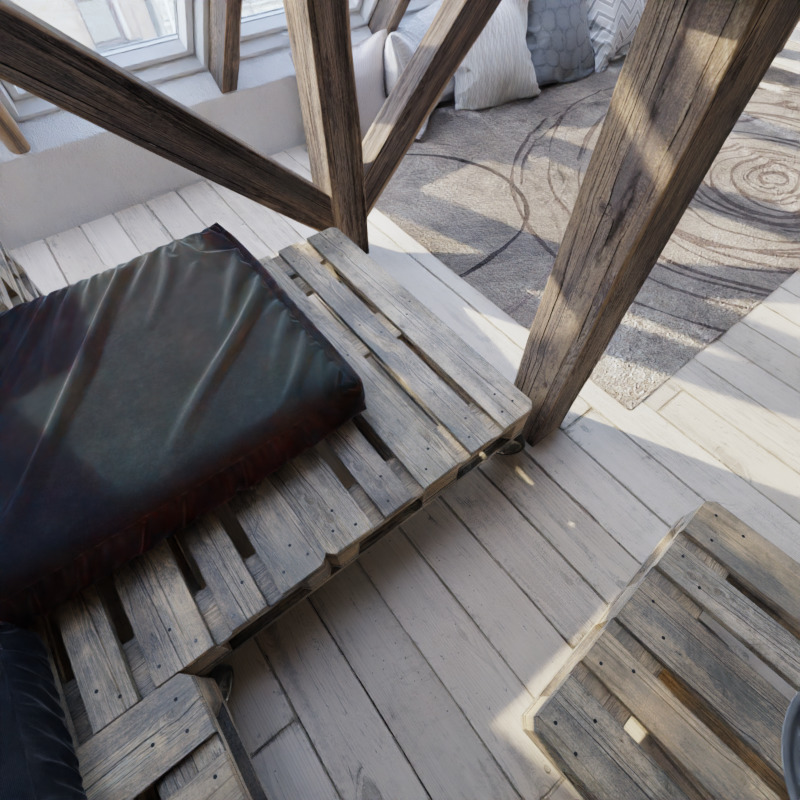} &
		\includegraphics[width=.19\linewidth]{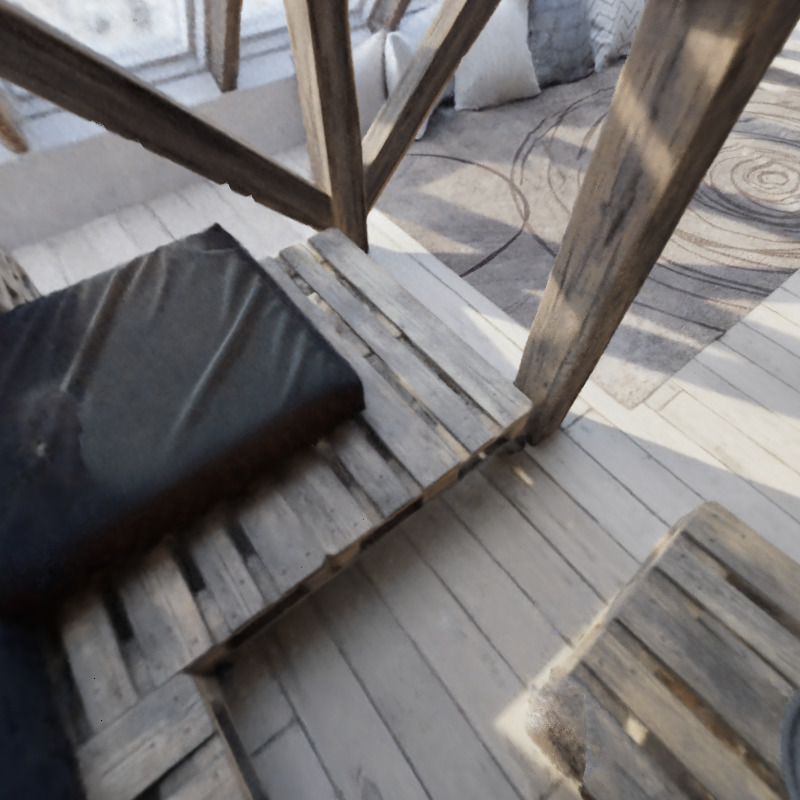} &
		\includegraphics[width=.19\linewidth]{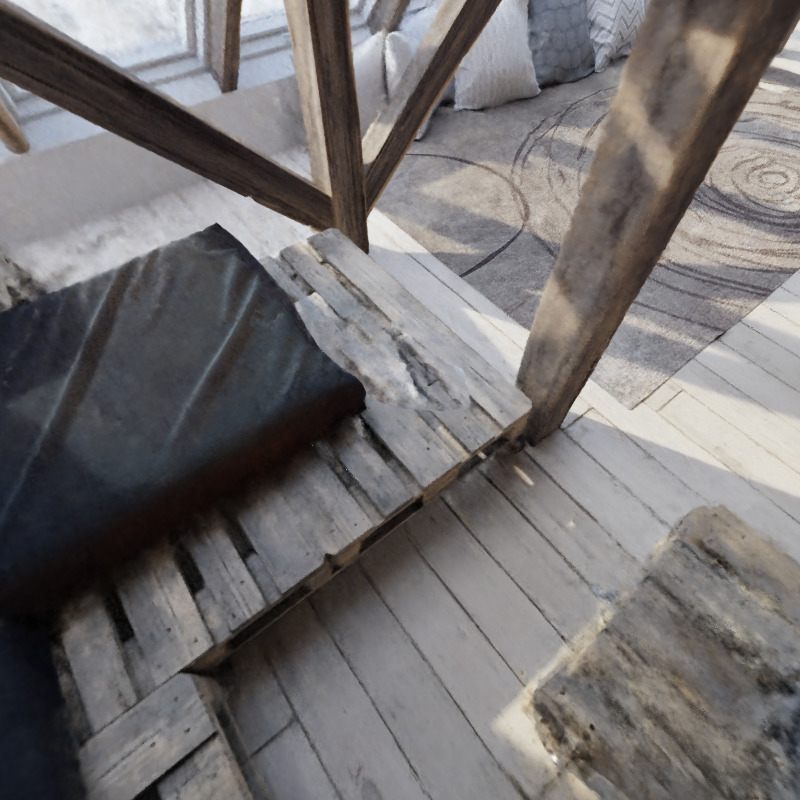} &
		\includegraphics[width=.19\linewidth]{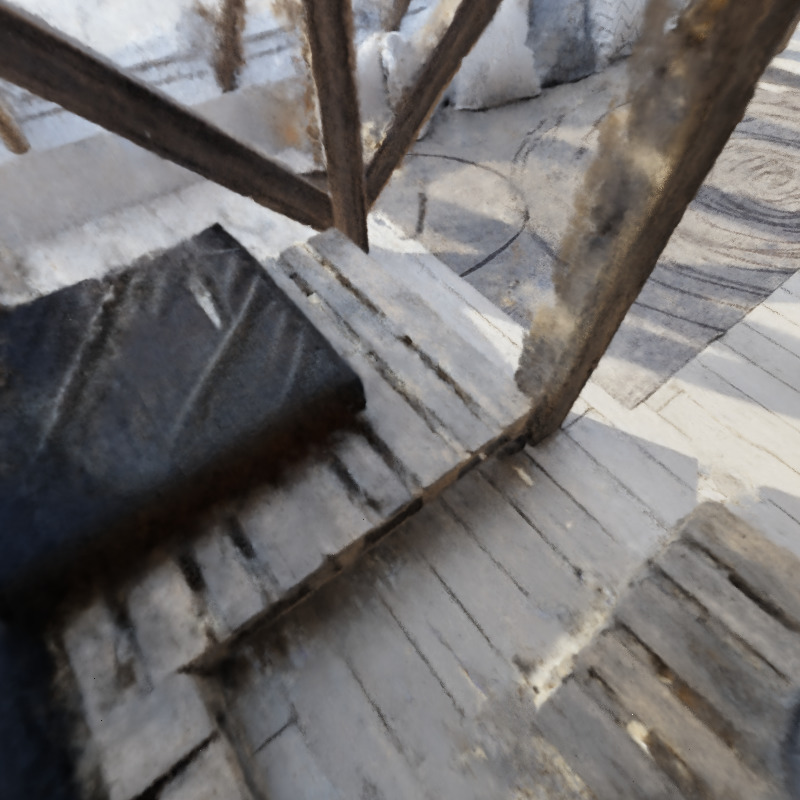} &
		\includegraphics[width=.19\linewidth]{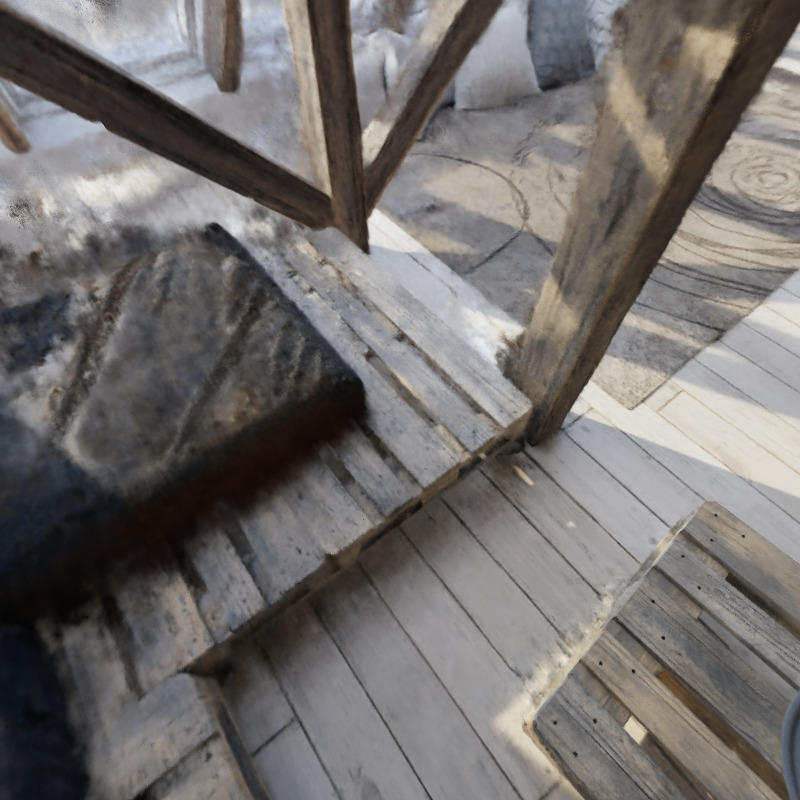} \\
		\includegraphics[width=.19\linewidth]{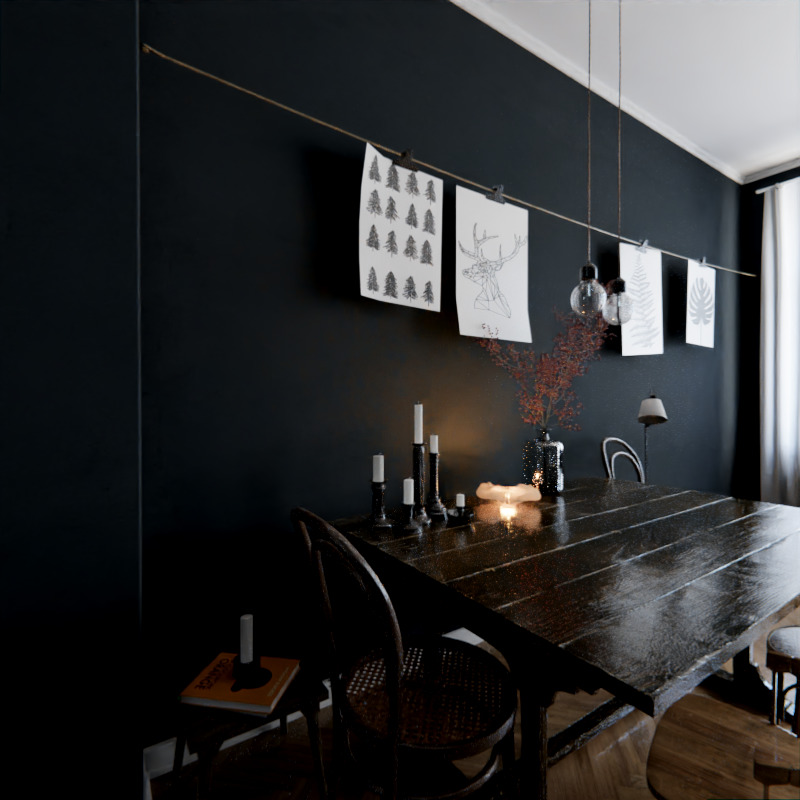} &
		\includegraphics[width=.19\linewidth]{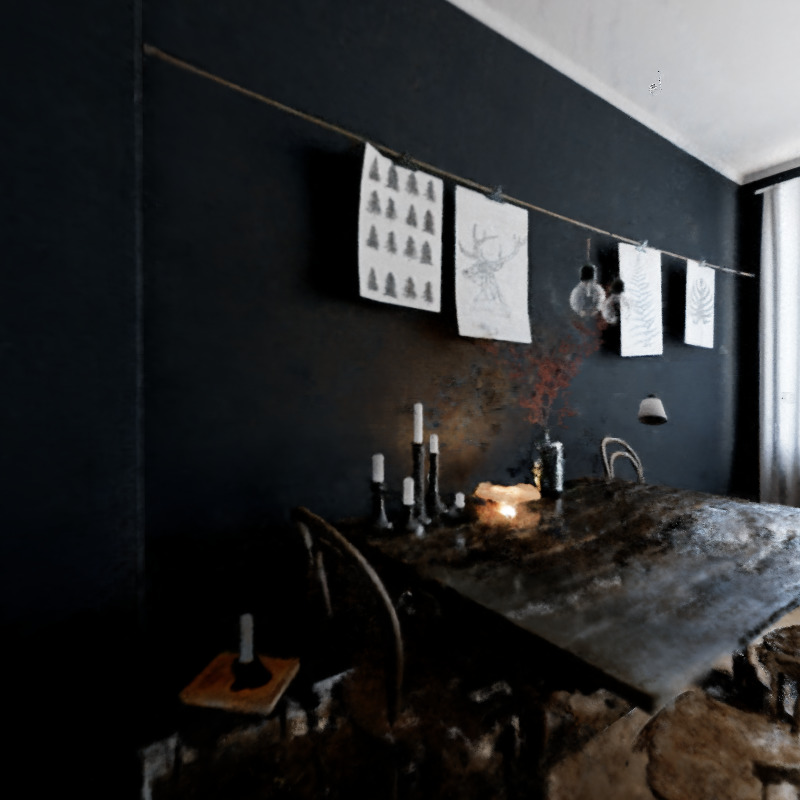} &
		\includegraphics[width=.19\linewidth]{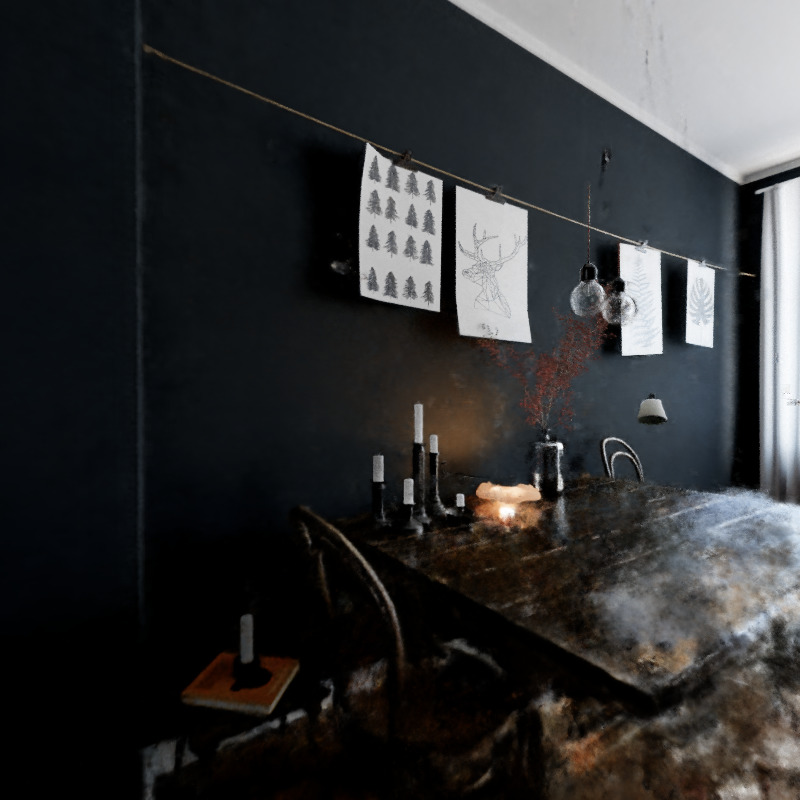} &
		\includegraphics[width=.19\linewidth]{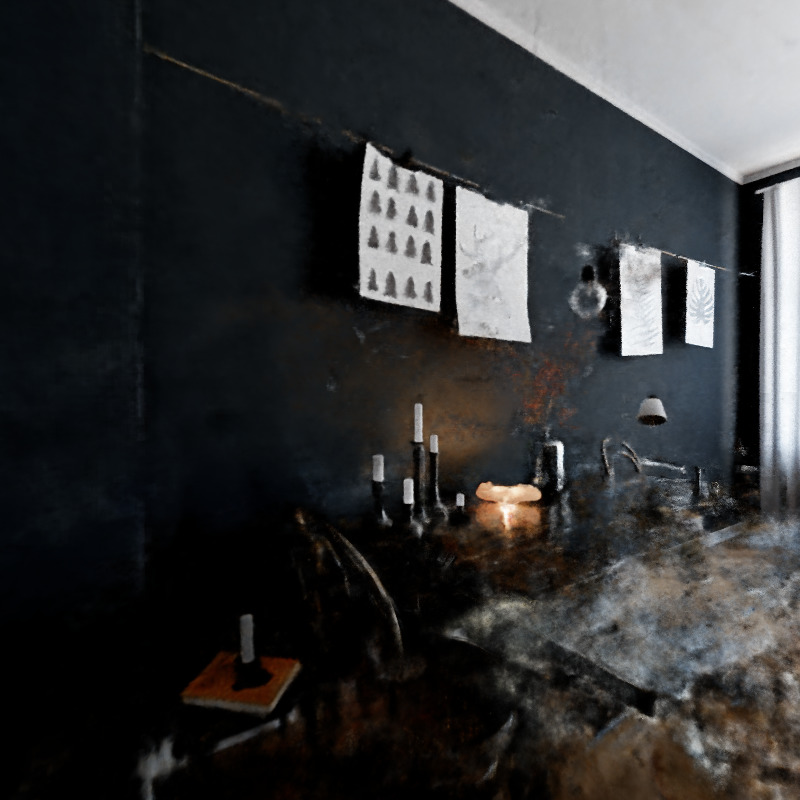} &
		\includegraphics[width=.19\linewidth]{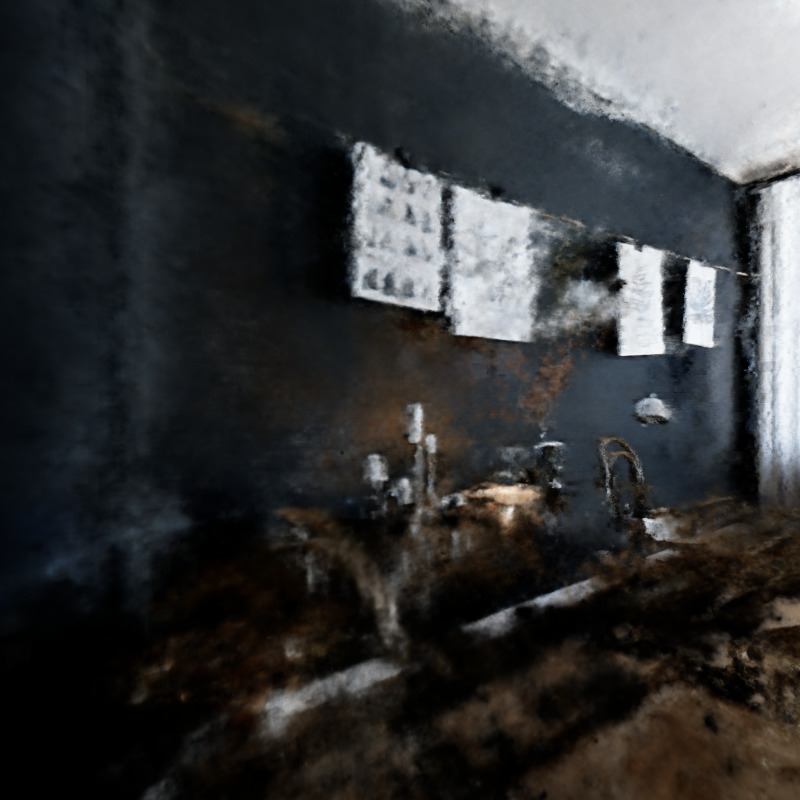} \\
		\includegraphics[width=.19\linewidth]{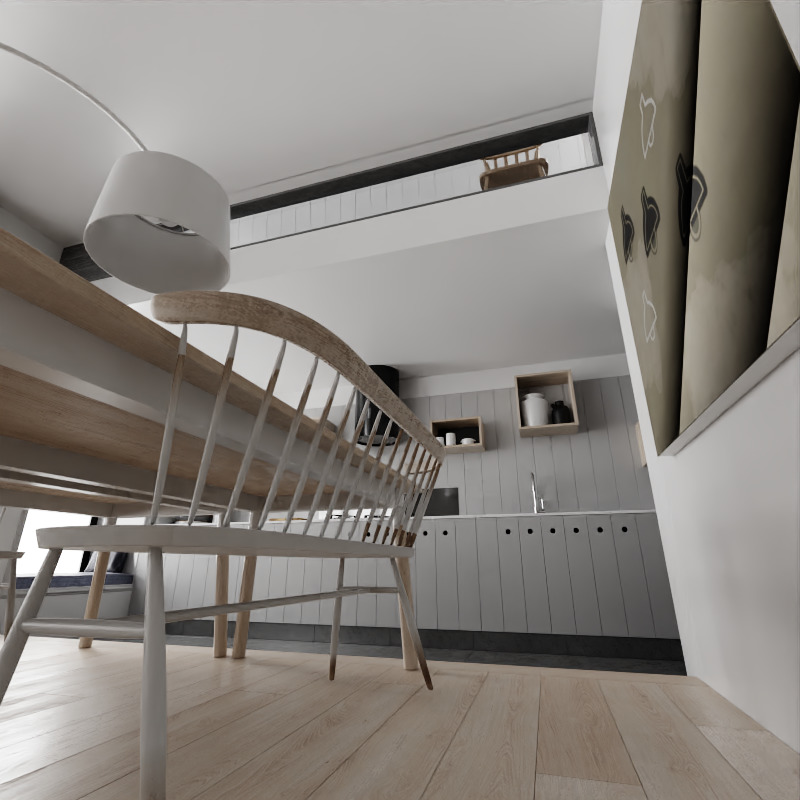} &
		\includegraphics[width=.19\linewidth]{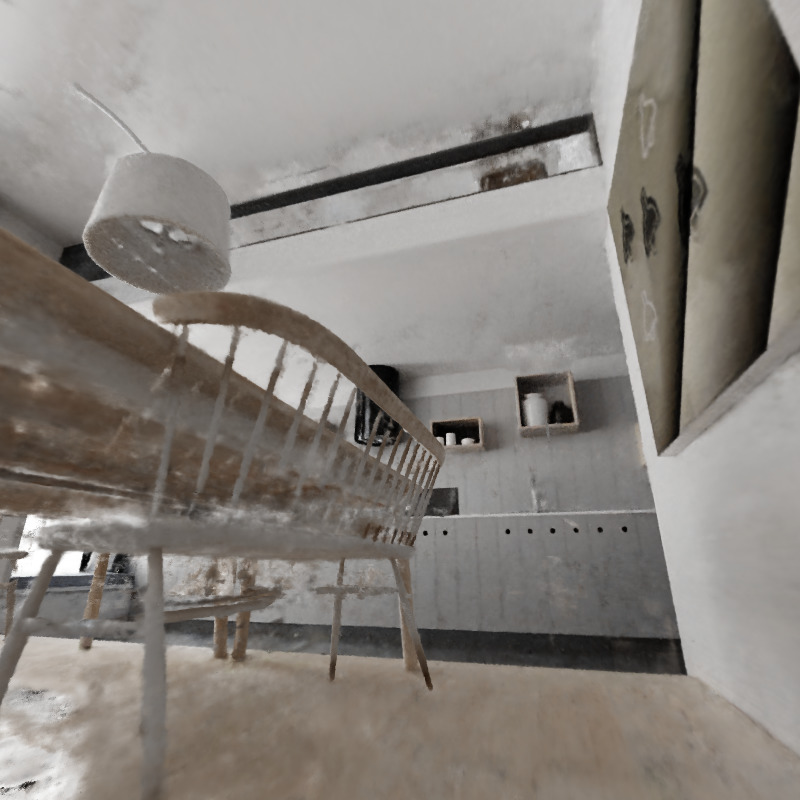} &
		\includegraphics[width=.19\linewidth]{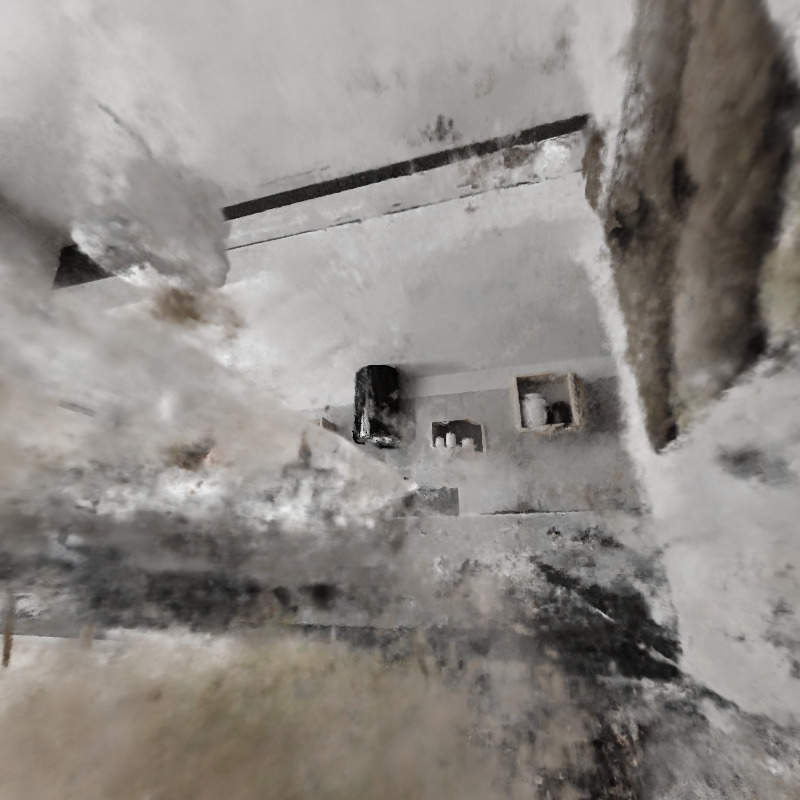} &
		\includegraphics[width=.19\linewidth]{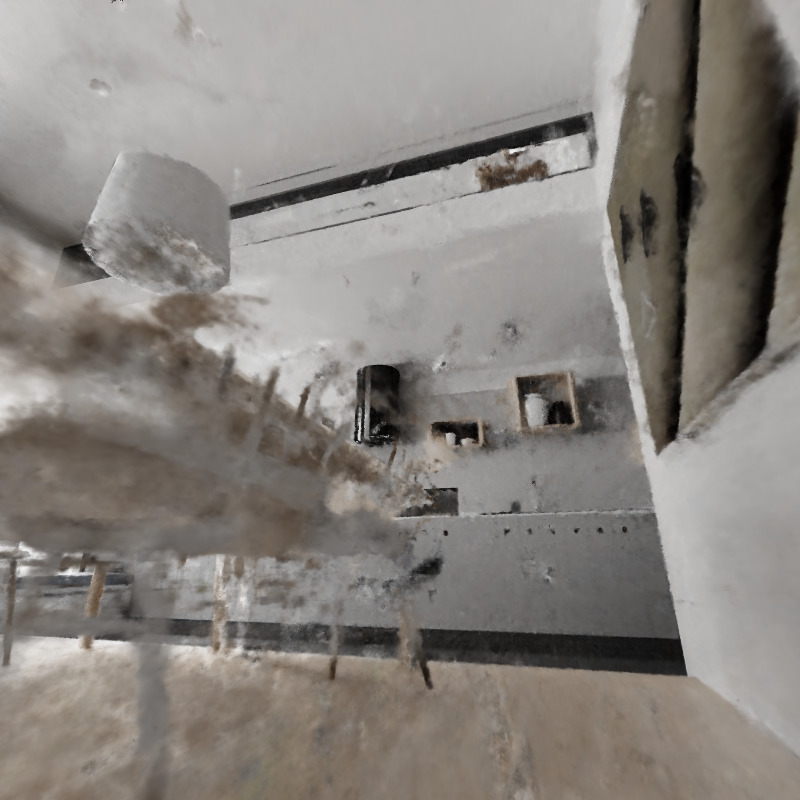} &
		\includegraphics[width=.19\linewidth]{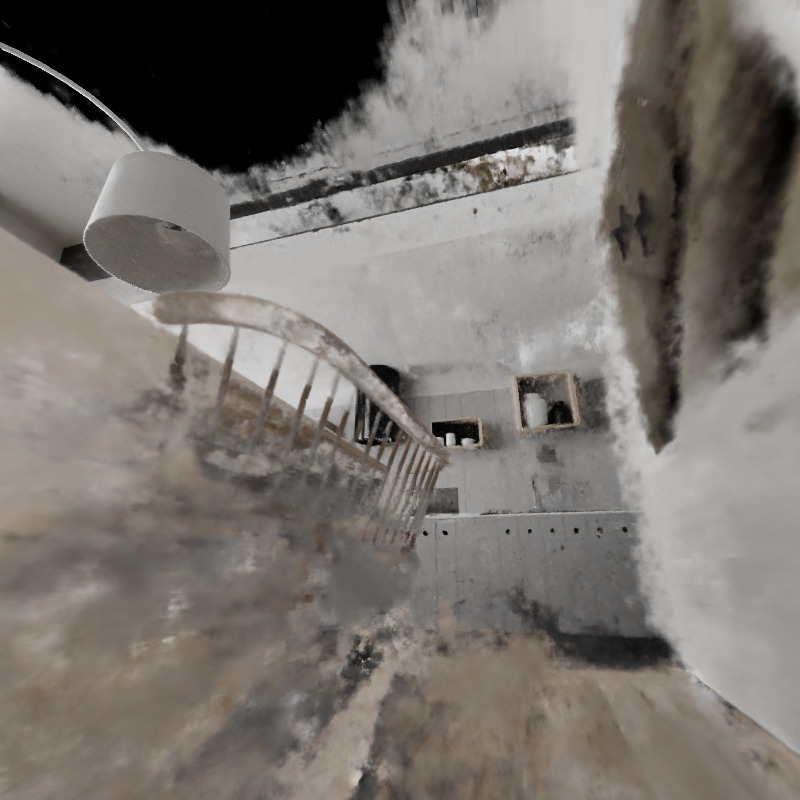} \\
		\includegraphics[width=.19\linewidth]{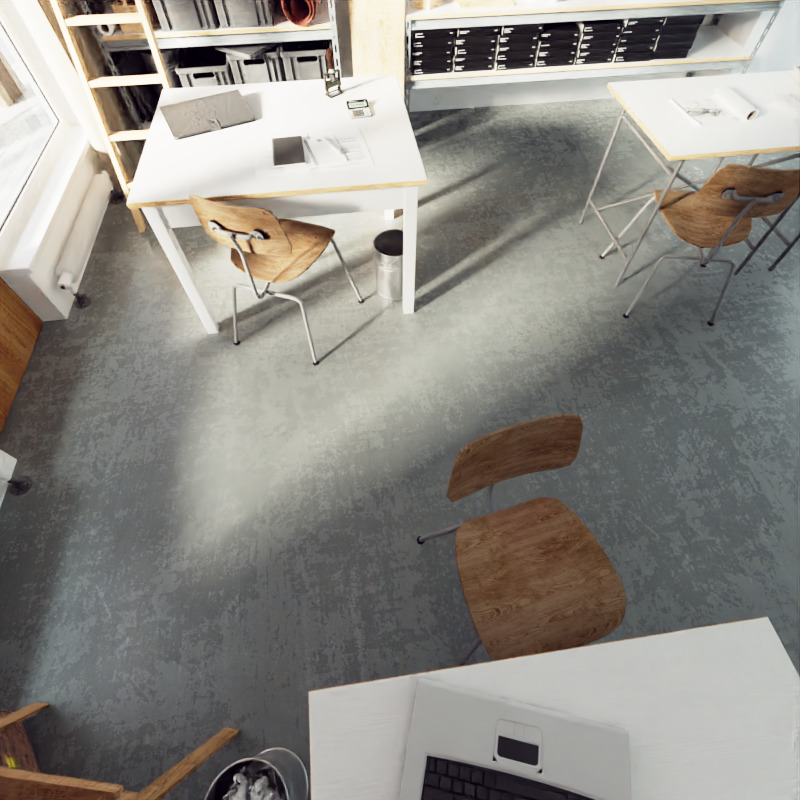} &
		\includegraphics[width=.19\linewidth]{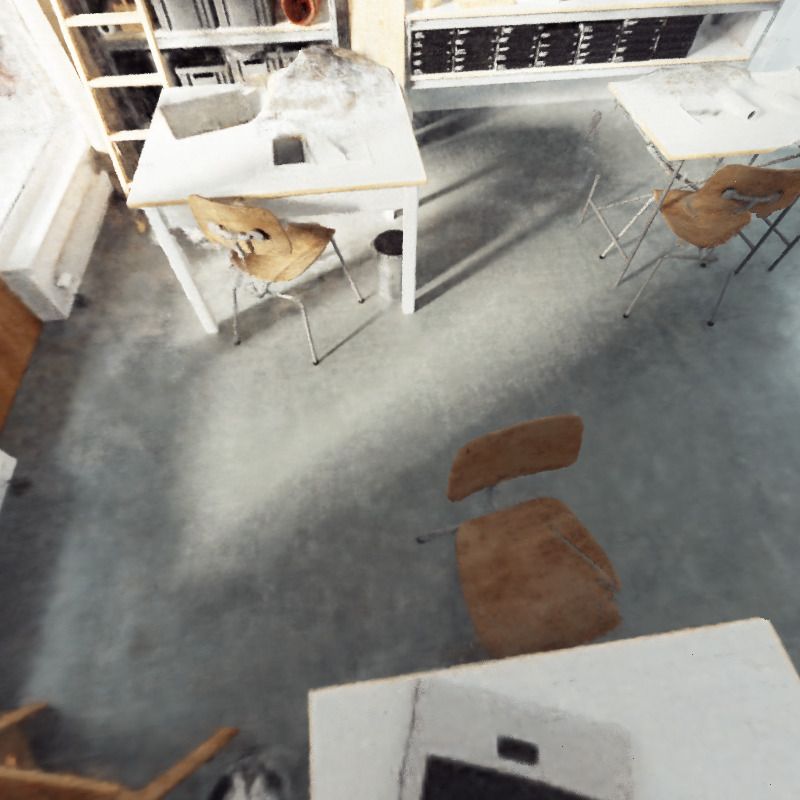} &
		\includegraphics[width=.19\linewidth]{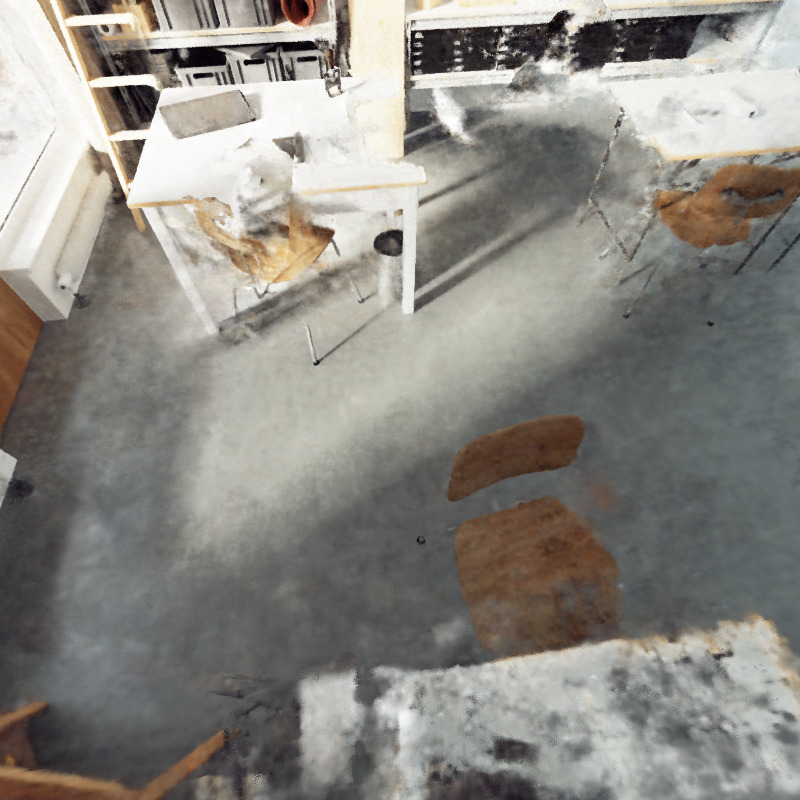} &
		\includegraphics[width=.19\linewidth]{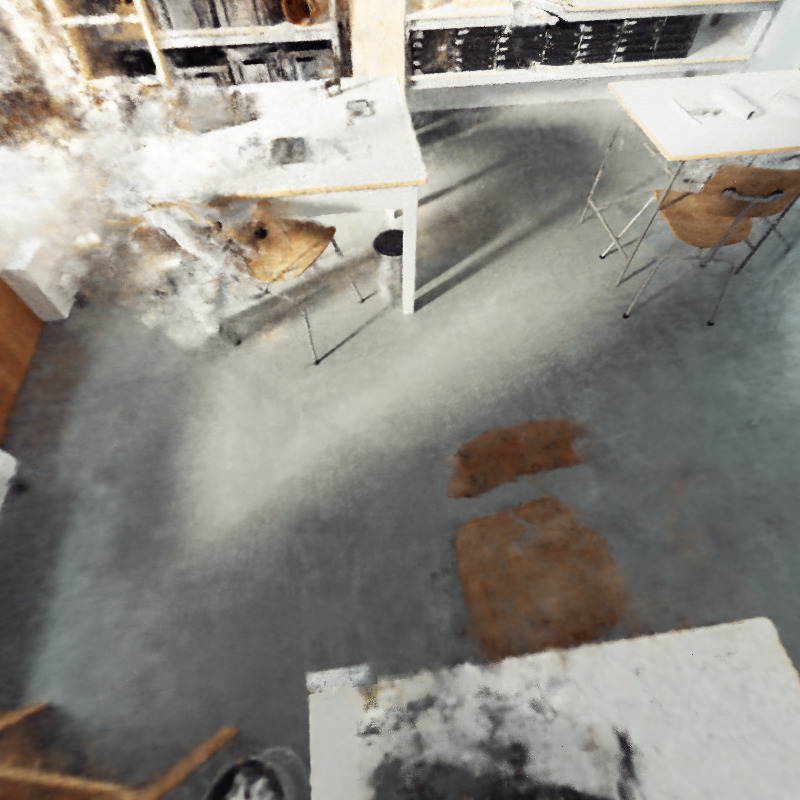} &
		\includegraphics[width=.19\linewidth]{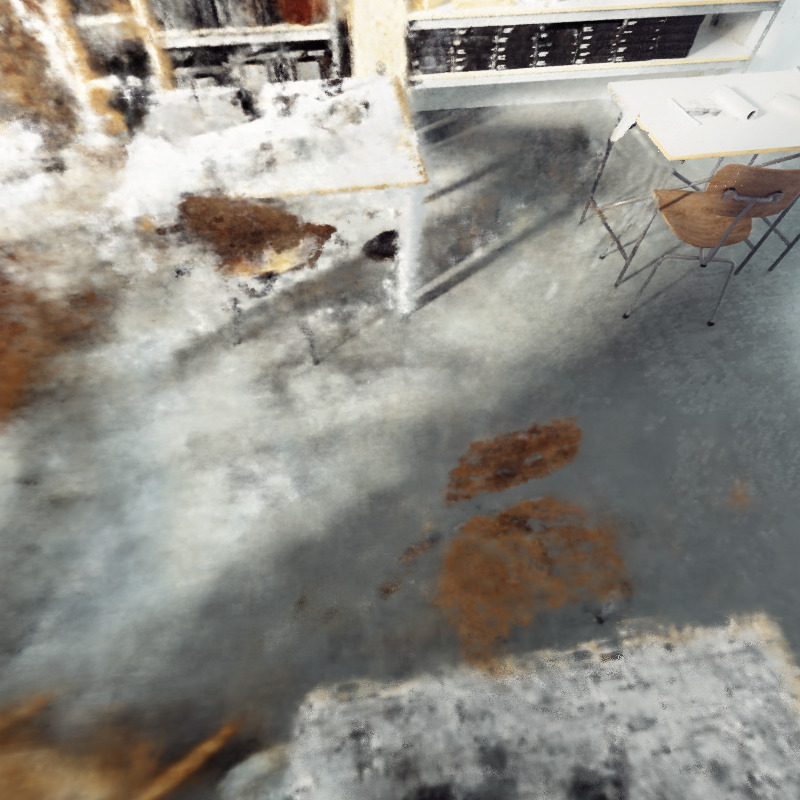} \\				
		\includegraphics[width=.19\linewidth]{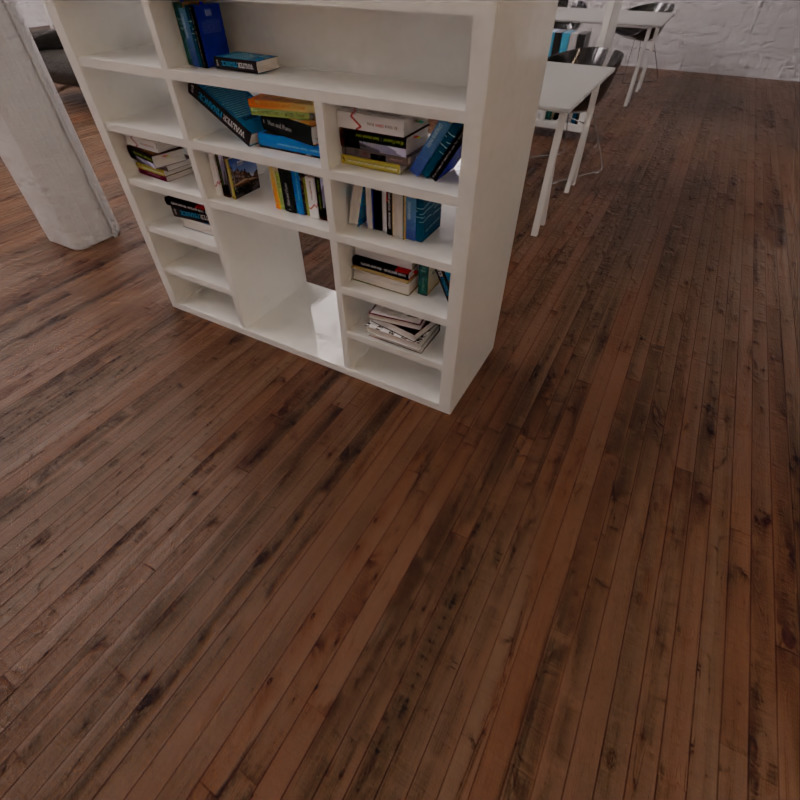} &
		\includegraphics[width=.19\linewidth]{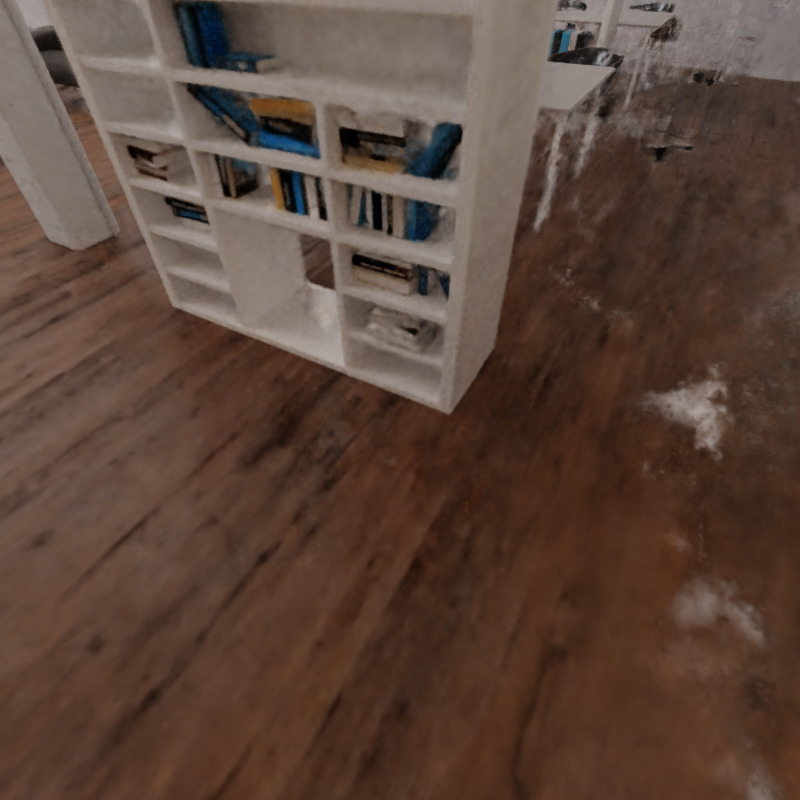} &
		\includegraphics[width=.19\linewidth]{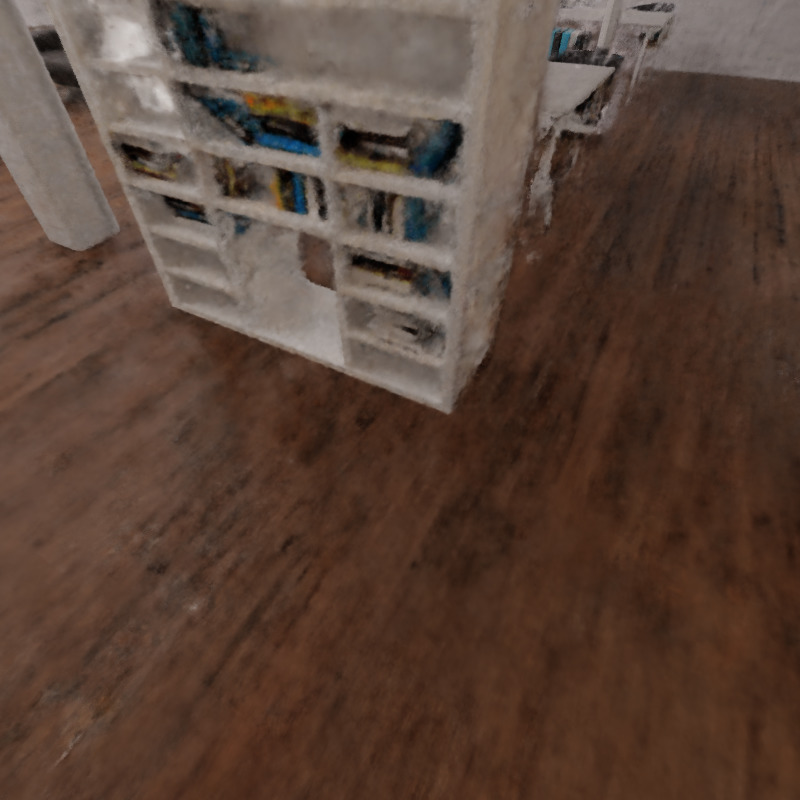} &
		\includegraphics[width=.19\linewidth]{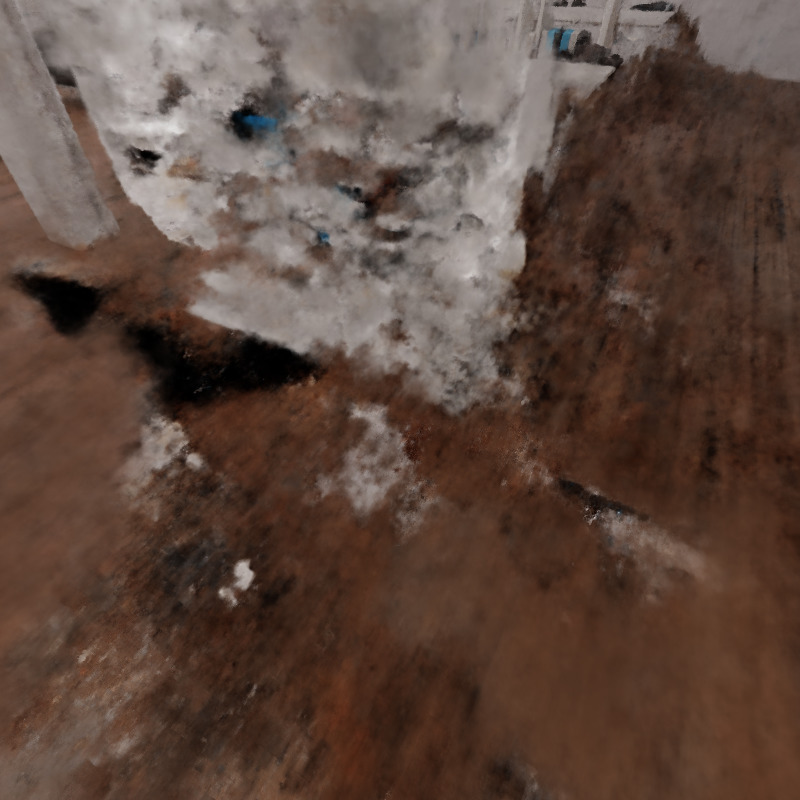} &
		\includegraphics[width=.19\linewidth]{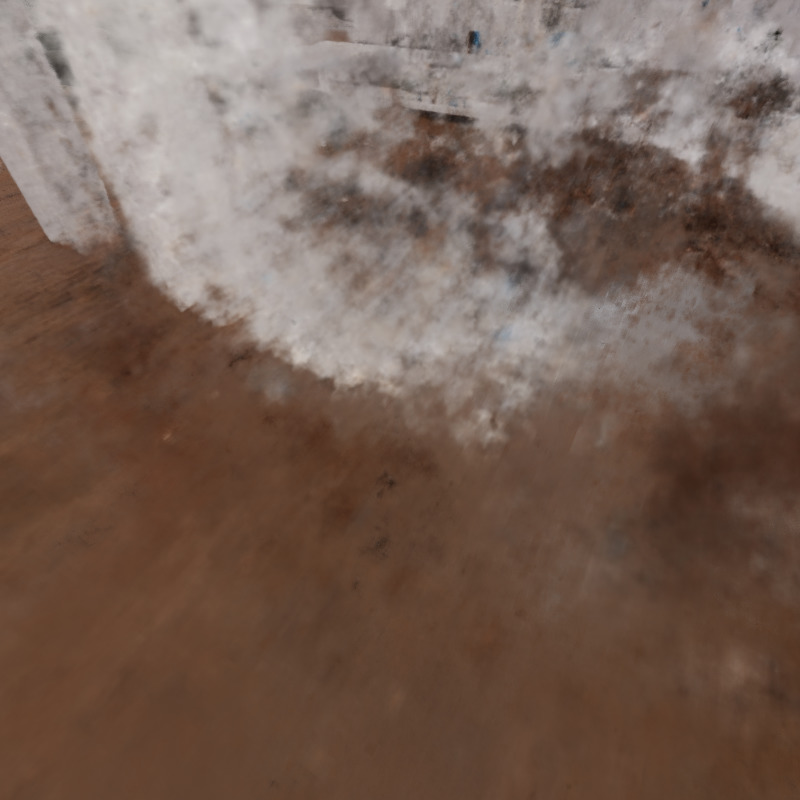} 
		
	\end{tabular}
	\caption{ 
	\label{fig:comparisons_random}
Images from the random camera sampling test set. We present a visual comparison to baselines (\textsc{Random},\textsc{Hemisphere}) and ActiveNerf. The first column shows the ground truth image. The scenes shown are \textsc{Livingroom1}, \textsc{Livingroom2}, \textsc{kitchen5}, \textsc{office6} and \textsc{office9}.
	}
\end{figure*}

We used 5 synthetic scenes modeled by professional artists to represent realistic indoor environments\footnote{The scenes are available for purchase at www.evermotion.com and are compatible with the Blender Cycles renderer.}. For each scene we construct a training set corresponding to each one of the algorithms we want to evaluate and multiple test sets that provide a good overview of the total quality throughout the scene. 

Our test-sets contain a total 150 views that are distinct from the training views. The test-sets are split in 3 sub-sets: 1) 50 random views using the \textsc{Hemisphere} capture style 2) 50 views using \textsc{Random} and 3) 50 views using our sampling process. The purpose of the multiple test sets is to evaluate each algorithm fairly throughout different camera distributions such that the quantitative metric evaluate the total quality throughout the scene. This avoids bias towards one of the aforementioned distributions, and allows a more comprehensive overall evaluation of our algorithm. We provide all the renderings for all views and all algorithms in our supplemental material. We also rendered free-view point paths which we provide in the supplemental video.

As we observe in Fig.~\ref{fig:comparisons_ours} and Fig.~\ref{fig:comparisons_random} the standard hemisphere captures of NeRF and MipNeRF fail to generalize in test sets coming from other distributions. Hemisphere views have a specific structure and objects that lie outside of the hemisphere are observed only from constrained angular directions and this 
%\TODO{WHAT DO YOU MEAN? 
allows for the NeRF model to overfit to the set of input cameras. While the random capture significantly outperforms the hemisphere capture in the generalized setting, we can still see significant artifacts because of the unstructured nature of the dataset. In theory, an infinite number of random views should allow for perfect reconstruction, but this is impractical since it is labor and computationally intensive. 

As we can see from the quantitative and qualititave evaluation~(Fig.~\ref{fig:comparisons_ours},\ref{fig:comparisons_random}), ActiveNerf~\cite{pan2022activenerf} does not always successfully choose the cameras that would allow for a good reconstruction. This happens for many reasons. First, the original NeRF models used has a hard time to converge in complicated scenarios that are not similar to the synthetic blender dataset, and it becomes even harder with the Bayesian model of Active-NeRF. Active-NeRF needs to get a notion of the scene to allow for good camera placements and in complicated scenes it can be challenging just from 20 initial cameras. Second, ActiveNeRF chooses cameras that maximize the uncertainty for the specific model they are training for, this does not guarantee that this uncertainty metric will generalize to other NeRF models and finally the memory and speed requirements do not allow for a huge number of candidate cameras similar to our method.

We can see that our capture style outperforms all other algorithms across the scenes and views both quantitatively in Tab.~\ref{tab:comparisons-per-scene} and qualitatively at Fig.~\ref{fig:comparisons_ours} and Fig.~\ref{fig:comparisons_random}. This supports our hypothesis that if we observe all parts of the scene while maintaining a uniform set of directions we will get an ideal reconstruction.

We also perform a visual analysis to provide insight on how different methods score against the energy function in Eq.~(\ref{eq:energy}). In Fig.~\ref{fig:floorplan_plot} we provide a visualization of the scores of each of the two terms of Eq.~(\ref{eq:energy}). We see that our method clearly observes all the nodes more often that the other baselines and we achieve better angular coverage for each node.

We also show in Fig.~\ref{fig:train_line} how our camera sampling improves in the test-sets as we introduce more and more cameras against the random cameras.

%\begin{table}[h]
%	\caption{
%		\label{tab:comparisons} {In this table we show the average numbers across the whole dataset for our scenes.}
%	}
%	{
%		\begin{tabular}{l|ccc|}
%			
%			 & $SSIM^\uparrow$   & $PSNR^\uparrow$    & $LPIPS^\downarrow$ \\
%			\hline 
%			Hsphere & & & \\
%			Hsphere2 & & & \\
%			Random & & & \\
%			ActiveNerf  & & & \\
%			Ours & & & \\
%		\end{tabular}
%	}
%\end{table}

\begin{figure}[!h]
	\includegraphics[width=\linewidth]{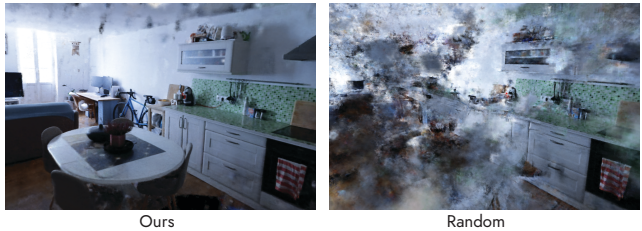}
	\vspace*{-5mm}\caption{ We also do a prelimary simulated evaluation of our method with a real scene, in which we captured approximately 1300 images, from which we exclude every 14th image and create a test-set. We use our algorithm to select the best 200 and we compared it against choosing 200 images at random. On average Our selection scored  16.4 PSNR while the random selection scored 13.8 PSNR.}
	\label{fig:real_scene}
\end{figure}

\subsection{Preliminary Real Scene Evaluation}

As discussed earlier (Sec.~\ref{sec:use-case}, we leave the actual integration of our method into a full capture system as future work.
Such a system would require either a user interface (e.g., on a phone) or interfacing with a robotic capture system (e.g., a drone).
However, it is instructive to see how well our method works on real data, so we present a very preliminary test on a real scene.
Since we are lacking a capture agent, we instead \emph{simulate} the ability to select new cameras as a proof of concept. Specifically, we achieve this by taking approximately 1300 photos (removing every 14th image to create a left-out test set), simulating ``random coverage'' of the scene. We then use our algorithm to select 200 cameras from this pool of randomly distributed images. 
This is evidently a very preliminary test, but the results shown in Fig.~\ref{fig:real_scene} show that our method performs significantly better than random selection.

%\begin{table}[h]
%	\caption{
%		\label{tab:comparisons} {In this table we show the average numbers across the whole dataset for our scenes.}
%	}
%	{
%		\begin{tabular}{l|ccc|}
%			
%			 & $SSIM^\uparrow$   & $PSNR^\uparrow$    & $LPIPS^\downarrow$ \\
%			\hline 
%			Hsphere & & & \\
%			Hsphere2 & & & \\
%			Random & & & \\
%			ActiveNerf  & & & \\
%			Ours & & & \\
%		\end{tabular}
%	}
%\end{table}

%-------------------------------------------------------------------------
\section{Conclusions}

We presented an efficient method for selecting cameras for NeRF capture in complicated environments, targeting free-viewpoint navigation. Our key contributions are the introduction of the angular and coverage metrics, and our fast optimization to propose the next best camera for NeRF reconstruction. Our method outperforms baselines and one previous method in overall perfomance; it is also faster than other methods and without significant overhead over baseline methods. An important attribute of our solution is that it is easily interpretable and can provide meaningful guidance and understanding to users without requiring additional images. One other benefit from the simplicity of our methods is that it could be adapted to vary the importance of the scene spatially; we leave this as future work.

Our method is not without limitations. One issue is that we have not investigated if our sampling is biased. If this is the case, no matter how many cameras we sample, we might not reach a ``perfect'' reconstruction and visual quality. Also, even though the method is efficient, it would benefit from even faster performance allow truly interactive capture. %,  in camera ``burst mode'' of 4-5 FPS.

There are numerous possibilities for future work. From a theoretical perspective, we are interested in studying other metrics of reconstruction quality in a more extensive and complete fashion. We are also very excited about the idea of integrating our approach in a mixed Augmented/Virtual Reality (AR/VR) context: for example we can guide an on-site (AR) user to take photos of a scene so that the remote VR user can very quickly be immersed in the same environment. Using our method in the context of drone capture would allow NeRF captures to be performed with high quality with little human intervention, rendering the approach much more useful and easy-to-use. 

%Also while we did not implement some features we believe that they are trivially incorporated to our method. One interesting extension is introducing occlusion aware energy terms that take into consideration the visibility of the nodes a camera observes. Another extension is to also enforce a uniform sampling of the distance from which we observe each node.
%\TODO{TODO: Ending with a positive note.}

%-------------------------------------------------------------------------
% bibtex
\bibliographystyle{eg-alpha-doi}  
\bibliography{egbibsample}        
%t
% biblatex with biber
%\printbibliography                

%-------------------------------------------------------------------------
\newpage

\end{document}